\documentclass[journal]{IEEEtran}
\ifCLASSINFOpdf
\else
\fi

\usepackage{amsmath,amssymb,amsfonts}
\usepackage{subcaption}
\usepackage{booktabs}
\usepackage{multirow}
\usepackage{colortbl} 
\usepackage{arydshln}
\usepackage{graphicx}
\usepackage{amsmath}
\usepackage{amssymb}
\usepackage{amsmath,amssymb,amsfonts}
\usepackage{algorithm}
\usepackage{booktabs}
\usepackage{algorithmic}
\usepackage{bm}
\usepackage{multirow}
\usepackage{amssymb}
\usepackage{subcaption}
\usepackage{mathtools}
\usepackage{makecell}
\usepackage{enumitem}
\usepackage{caption}
\usepackage{graphicx}
\usepackage{amsmath}
\usepackage{amsthm}
\usepackage{booktabs}
\usepackage{algorithm}
\usepackage{algorithmic}
\usepackage{graphicx}
\usepackage{amsmath}
\usepackage{amsmath,amssymb,amsfonts}
\usepackage{algorithm}
\usepackage{booktabs}
\usepackage{algorithmic}
\usepackage{bm}
\usepackage{multirow}
\usepackage{amssymb}
\usepackage{color}
\usepackage{subcaption}
\usepackage{multirow}
\usepackage{amssymb}
\usepackage{subcaption}
\usepackage{mathtools}
\usepackage{makecell}
\usepackage{enumitem}
\usepackage{caption}
\begin{document}
%
\title{Knowledge Distillation Using Hierarchical Self-Supervision Augmented Distribution}
%
%
%

\author{Chuanguang~Yang,
        Zhulin~An,
        Linhang~Cai,
        and~Yongjun~Xu
\thanks{Chuanguang Yang and Linhang Cai are with Institute of Computing Technology, Chinese Academy of Sciences, Beijing 100190, China and University of Chinese Academy of Sciences, Beijing 100049, China. Email: \{yangchuanguang, cailinhang19g\}@ict.ac.cn}
\thanks{Zhulin An and Yongjun Xu are with Institute of Computing Technology, Chinese Academy of Sciences, Beijing 100190, China. Email: \{anzhulin, xyj\}@ict.ac.cn}
\thanks{Zhulin An is the corresponding author.}
\thanks{Manuscript received July 31, 2021; revised May 19, 2022; accepted
	June 20, 2022.}}

%
%

\markboth{IEEE TRANSACTIONS ON NEURAL NETWORKS AND LEARNING SYSTEMS}%
{Shell \MakeLowercase{\textit{et al.}}: Bare Demo of IEEEtran.cls for IEEE Journals}
%



\maketitle

\begin{abstract}
Knowledge distillation (KD) is an effective framework that aims to transfer meaningful information from a large teacher to a smaller student.  Generally, KD often involves how to define and transfer knowledge. Previous KD methods often focus on mining various forms of knowledge, for example, feature maps and refined information. However, the knowledge is derived from the primary supervised task and thus is highly task-specific. Motivated by the recent success of self-supervised representation learning, we propose an auxiliary self-supervision augmented task to guide networks to learn more meaningful features. Therefore, we can derive soft self-supervision augmented distributions as richer dark knowledge from this task for KD. Unlike previous knowledge, this distribution encodes joint knowledge from supervised and self-supervised feature learning. Beyond knowledge exploration, we propose to append several auxiliary branches at various hidden layers, to fully take advantage of hierarchical feature maps. Each auxiliary branch is guided to learn self-supervision augmented task and distill this distribution from teacher to student. Overall, we call our KD method as Hierarchical Self-Supervision Augmented Knowledge Distillation (HSSAKD). Experiments on standard image classification show that both offline and online HSSAKD achieves state-of-the-art performance in the field of KD. Further transfer experiments on object detection further verify that HSSAKD can guide the network to learn better features. The code is available at https://github.com/winycg/HSAKD.

\end{abstract}


%
\IEEEpeerreviewmaketitle

\section{Introduction}
\IEEEPARstart{A}LTHOUGH deep CNNs have achieved great successes over a series of visual tasks~\cite{he2016deep,ren2016faster,zhao2019object,yang2020gated}, it is often hard to deploy over-parameterized networks on resource-limited edge devices. Typical solutions can be efficient architecture design~\cite{zhang2018shufflenet,sandler2018mobilenetv2,ma2018shufflenet,yang2020gated,ding2021bnas}, model pruning~\cite{yang2019multi,lin2021filter,cai2022prior} and knowledge distillation~\cite{hinton2015distilling,ge2018low,ge2019distilling,zhang2021ekd,yang2022rd,yang2022multi}. Knowledge Distillation (KD) provides a simple framework that transfers knowledge from a pre-trained high-capacity teacher network to a smaller and faster student network. Compared with independent training, the student network can generalize better benefiting from the additional guidance. The current pattern of KD can be summarized as two critical aspects: (1) what kind of knowledge encapsulated in teacher network can be explored for KD; (2) How to effectively transfer knowledge from teacher to student.

\begin{figure}
	\centering 
	\begin{subfigure}[t]{0.45\textwidth}
		\centering
		\includegraphics[width=\textwidth]{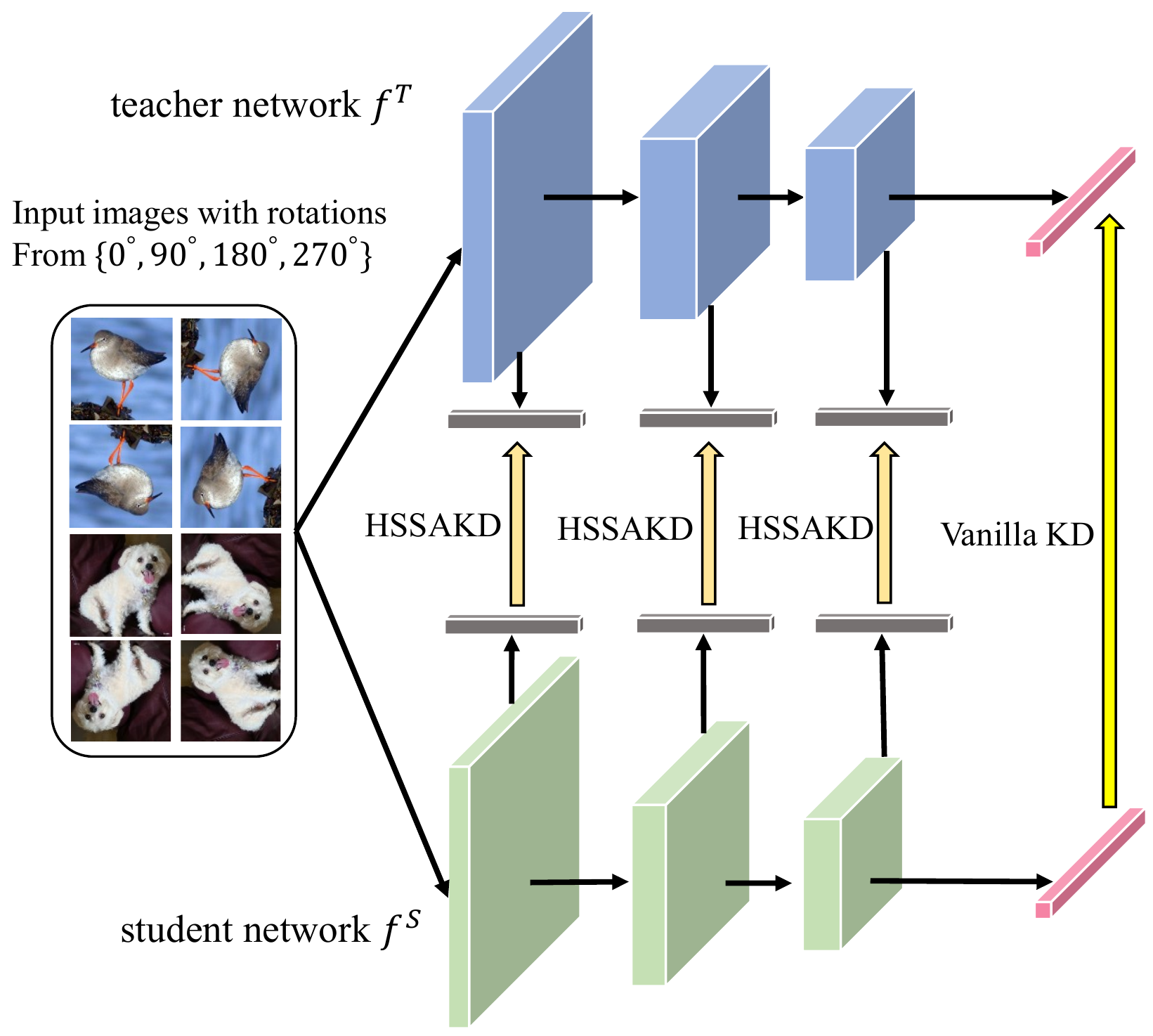}
		\caption{Hierarchical Self-Supervision Augmented KD}
		\label{intro1}
	\end{subfigure}
	\begin{subfigure}[t]{0.47\textwidth}
		\centering
		\includegraphics[width=\textwidth]{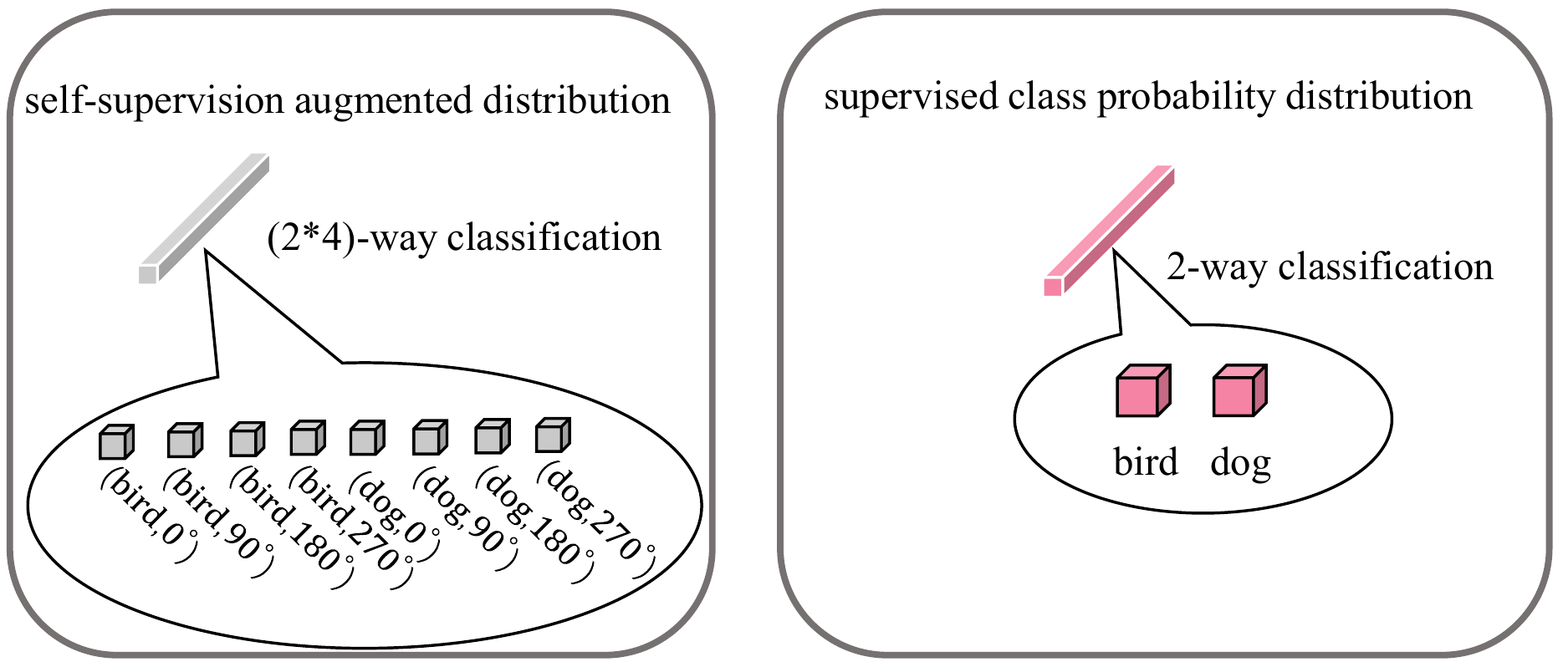}
		\caption{Legend of distilled knowledge types.}
		\label{intro2}
	\end{subfigure}
	
	\caption{An overview of our proposed HSSAKD and distilled knowledge types.}
	\label{intro}   
\end{figure}

The original KD~\cite{hinton2015distilling} minimizes the KL-divergence of predictive class probability distributions between student and teacher networks, which makes intuitive sense to force the student to mimic how a superior teacher generates the final predictions. However, such highly abstract dark knowledge ignores much comprehensive information encoded in hidden layers. Later works naturally proposed to transfer feature maps~\cite{romero2014fitnets} or their refined information~\cite{zagoruyko2016paying,heo2019knowledge,Sungsoo19Variational} between intermediate layers of teacher and student. A reasonable interpretation of the success of feature-based distillation lies in that hierarchical feature maps throughout the CNN represent the intermediate learning process with inductive biases of the final solution. Beyond knowledge alignment limited in the individual sample, more recent works~\cite{peng2019correlation,tian2019distillation} leverage cross-sample correlations or dependencies in high-layer feature embedding space. A common characteristic of previous works often mines knowledge only derived from a single supervised classification task. The task-specific knowledge may hinder the network from exploring richer feature representations. 

To excavate more meaningful knowledge, it makes sense to introduce an auxiliary task to guide the network to produce extra representational information, complementary to the primary supervised classification knowledge. From the perspective of representation learning, a natural method is to consider an extra self-supervised task. The idea behind self-supervised learning is to learn \emph{general feature representations} guided by artificial supervisions derived from data itself via a pre-defined pretext task. Generally, learned features could well generalize to other semantic tasks, \emph{e.g.} image classification~\cite{noroozi2016unsupervised,gidaris2018unsupervised}. Some self-supervised learning works can also be seen as a classification problem with artificial label space, \emph{e.g.} random rotations~\cite{noroozi2016unsupervised} and jigsaw puzzles with shuffled permutations~\cite{gidaris2018unsupervised}. By forcing a network to recognize which transformation is applied on an input transformed image, the network would learn general semantic features. 

Based on the fact that both supervised and self-supervised tasks are classification problems, we propose to combine two label spaces via a Cartesian product and model full-connected interactions, inspired by~\cite{lee2020self}. Formally, given a $N$-way supervised task and a $M$-way self-supervised task, we merge them into a $N*M$-way joint task, where each label is a 2-tuple that consists of one supervised label and one self-supervised label. We name this joint task as \textbf{\emph{self-supervision augmented task}} due to that supervised classification is our primary task. Based on this task, we can naturally derive a joint class probability distribution as a new form of dark knowledge. For ease of notation, we name this knowledge as \textbf{\emph{self-supervised augmented distributions}}. Fig.~\ref{intro} illustrates a simple example of the joint label space. Through learning and distilling self-supervision augmented knowledge, the network can benefit from supervised and self-supervised representation learning jointly.

Another valuable problem lies in how to effectively learn and transfer the probabilistic knowledge between teacher and student. Response-based KD~\cite{hinton2015distilling} aligns probability distributions only in the final layer, because it may be difficult to explicitly derive comprehensive probability distributions from hidden feature maps over the original architecture. Inspired by deeply-supervised nets (DSN)~\cite{lee2015deeply}, a natural idea is to append several auxiliary branches to a network at various hidden layers. As illustrated in Fig.~\ref{intro}, those branches enable the network to generate multi-level self-supervision augmented distributions from hierarchical feature maps, which further allows us to perform one-to-one distillation in hidden layers towards probabilistic knowledge. However, designing a suitable auxiliary branch to learn an auxiliary task is also a non-trivial problem. Unlike a simple linear classifier in DSN, our auxiliary branch includes an extra feature extraction module. The module is responsible for transforming the feature map derived from the backbone network to adapt for a new task.

Given a single network, we guide the backbone network to learn the primary supervised task and all auxiliary branches to learn our self-supervision augmented task. Surprisingly, this method significantly improves the backbone network performance over the primary supervised task compared to the baseline, \emph{e.g.} an average accuracy gain of 4.15\% across widely used network architectures on CIFAR-100. We attribute the gain to our auxiliary self-supervision augmented task that can guide the backbone network to learn better feature representations. 

Based on the success of the auxiliary self-supervision augmented task, we think it is promising to distill self-supervision augmented distributions towards all auxiliary branches in a one-to-one manner. Because distributions at various depths are derived from hierarchical feature maps respectively, we name our method as \textbf{\emph{Hierarchical Self-Supervision Augmented Knowledge Distillation}} (HSSAKD). For offline HSSAKD, we first train a teacher network using the self-supervision augmented task and then transfer self-supervision augmented distributions as soft labels to supervise a student network. For online HSSAKD, we perform mutual distillation of self-supervision augmented distributions among several networks jointly trained from scratch. Note that all auxiliary branches are only used to assist knowledge transfer and would be dropped at the inference period, leading to no extra costs compared with the original network.  

Experimental results on standard CIFAR-100 and ImageNet demonstrate that both offline and online HSSAKD can significantly achieve better performance than other competitive KD methods. Fig.~\ref{intro} illustrates the superiority over SOTA CRD~\cite{tian2019distillation} and SSKD~\cite{DBLP:conf/eccv/XuLLL20} with large accuracy gains. Extensive results on transfer experiments to classification and detection suggest that HSSAKD can teach a network to learn better general features for semantic recognition tasks.

This paper is an extension of our conference version~\cite{yang2021hierarchical}. The new contributions are summarized as follows:
\begin{itemize}[noitemsep,nolistsep,,topsep=0pt,parsep=0pt,partopsep=0pt]
	\item We examine the effectiveness of several variants involving supervised as well as self-supervised tasks and their derived knowledge for training and KD. We demonstrate that our self-supervision augmented task is a powerful auxiliary task for representation learning. 
	\item Based on the success of offline HSSAKD, we extend it to online HSSAKD for distilling self-supervision augmented distributions mutually. To our knowledge, this method may be the first work that utilizes self-supervised representation knowledge for online KD.
	\item Our online HSSAKD also achieves SOTA performance on standard image classification benchmarks and significantly surpasses other competitive online KD methods.
	\item More systematical algorithm descriptions, more detailed ablation study and analyses for offline and online HSSAKD are presented.
\end{itemize}
%
%
%
%

\section{Related Work}
\subsection{Offline Knowledge Distillation} The conventional offline KD conducts a two-stage training pipeline. It first trains a powerful teacher network and transfers knowledge from the pre-trained teacher to a smaller and faster student network via a meaningful distillation strategy.  The seminal KD~\cite{hinton2015distilling} popularizes the pattern of knowledge transfer with a soft class posterior probability distribution. Later methods further explore feature-based information encapsulated in hidden layers for KD, such as intermediate feature maps~\cite{romero2014fitnets}, attention maps~\cite{zagoruyko2016paying}, gram matrix~\cite{Gift17} and activation boundaries~\cite{heo2019knowledge}. Unlike exploiting features from hidden layers, more recent works model cross-sample high-order relationships using high-level abstract feature embeddings for KD. Typical metrics of relationships between various samples can be pairwise distance~\cite{park2019relational}, correlation congruence~\cite{peng2019correlation} and similarity graph~\cite{Tung2019Similarity}. Based on this idea, Ye \emph{et al.}~\cite{ye2020distilling} distill cross-task knowledge via high-order relationship matching. Unlike explicitly transferring relationships, CRD~\cite{tian2019distillation} applies contrastive learning among samples from both the teacher and student networks to implicitly perform contrastive representation distillation.

Inspired by recent self-supervised learning~\cite{chen2020simple}, SSKD~\cite{DBLP:conf/eccv/XuLLL20} extracts structured knowledge from an auxiliary self-supervised task. Although both SSKD and our HSSAKD take advantage of self-supervised learning, the difference lies in how to leverage self-supervision to define knowledge. SSKD models contrastive relationships by forcing various transformed versions with rotations over the same image closed.  As verified in~\cite{yang2021hierarchical}, this contrastive distribution may destroy the original visual semantics. In contrast, our HSSAKD explores a more informative knowledge form by encoding supervised and self-supervised distillation into a unified distribution. It does not interfere with a network in recognizing the original semantics, which is very different from SSKD.

Beyond knowledge mining, another aspect is to examine the transfer strategy. The conventional methods~\cite{romero2014fitnets,zagoruyko2016paying,Gift17} often use L2 loss to align feature maps. Moreover, VID~\cite{Sungsoo19Variational} maximizes the mutual information between matched features. Wang \emph{et al.}~\cite{wang2018adversarial} utilize the idea of generative adversarial network to make feature distributions similar. PKT~\cite{passalis2020probabilistic} uses different kernels to estimate the probability distribution along with adaptive divergence metrics for KD. Mirzadeh \emph{et al.}~\cite{mirzadeh2020improved} point out that a large capability gap between teacher and student may lead to a negative supervisory efficacy.  To bridge the gap, TAKD~\cite{mirzadeh2020improved} inserts an intermediate-sized network as a teacher assistant and performs multi-step KD. With the same motivation, Passalis \emph{et al.}\cite{passalis2020heterogeneous} introduce a proxy model to “explaining” knowledge teacher works to the student. However, extra teacher models increase the complexity of the training pipeline. We append several well-designed auxiliary branches to alleviate the knowledge gap. It facilitate comprehensive knowledge transfer from hierarchical feature maps.

\subsection{Online Knowledge Distillation} Unlike the conventional offline KD, online KD has not
a specific teacher-student role and trains multiple student networks simultaneously to learn from each other in a peer-teaching manner. DML~\cite{zhang2018deep} indicates that a cohort of student networks can benefit from transferring class probability distributions mutually in an online manner. CL-ILR~\cite{song2018collaborative} extends this framework to a hierarchical network with multiple classification heads. In contrast to a mutual mechanism, ONE~\cite{zhu2018knowledge} produces gated ensemble logits as a teacher role for online KD. Chen \emph{et al.}~\cite{chen2020online} point out the homogenization problem in ONE and utilize a self-attention module to compute aggregation weights for ensemble logits. With the same motivation, KDCL~\cite{guo2020online} proposes to adaptively aggregate logits via a principle manner. Beyond class probabilities, AFD~\cite{chung2020feature} performs mutual mimicry between feature distributions via an online adversarial training framework. Yang \emph{et al.}~\cite{yang2021multi,yang2022mcl} apply contrastive learning among multiple peer networks for online KD. Our HSSAKD can also be extended to online KD and perform mutual mimicry of hierarchical self-supervised augmented distributions. To our knowledge, HSSAKD may be the first work that utilizes self-supervised representation knowledge for online KD.

\subsection{Self-Supervised Learning} Self-supervised learning has been widely applied for computer vision, such as image classification~\cite{chen2020simple,misra2020self,bhat2021distill}, video recognition~\cite{yao2020video,luo2020video} and human pose estimation~\cite{zhang2021learning}. The basic idea of self-supervised learning is to design a pretext task that can guide the network to learn meaningful feature representations. Some seminal self-supervised learning works can be regarded as a classification framework that trains a network to learn which transformation is applied to a transformed input image. In this vein, typical transformations can be rotations~\cite{Spyros2018Unsupervised}, jigsaw~\cite{noroozi2016unsupervised} and colorization~\cite{zhang2016colorful}. Another vein is to use contrastive learning. More recently, SimCLR~\cite{chen2020simple} and PIRL~\cite{misra2020self} learn invariant feature representations by maximizing the consistency among various transformed versions of the same image. We remark that both SSKD and our HSSAKD  refer to self-supervised works to define knowledge. SSKD~\cite{DBLP:conf/eccv/XuLLL20} follows SimCLR~\cite{chen2020simple} to extract self-supervised contrastive-based relationships. In contrast, HSSAKD combines the supervised and self-supervised tasks as a joint classification task and aims to transfer richer joint knowledge.

\section{Methodology}

\begin{figure*}[tbp]  
	\centering  
	\includegraphics[width=1.\linewidth]{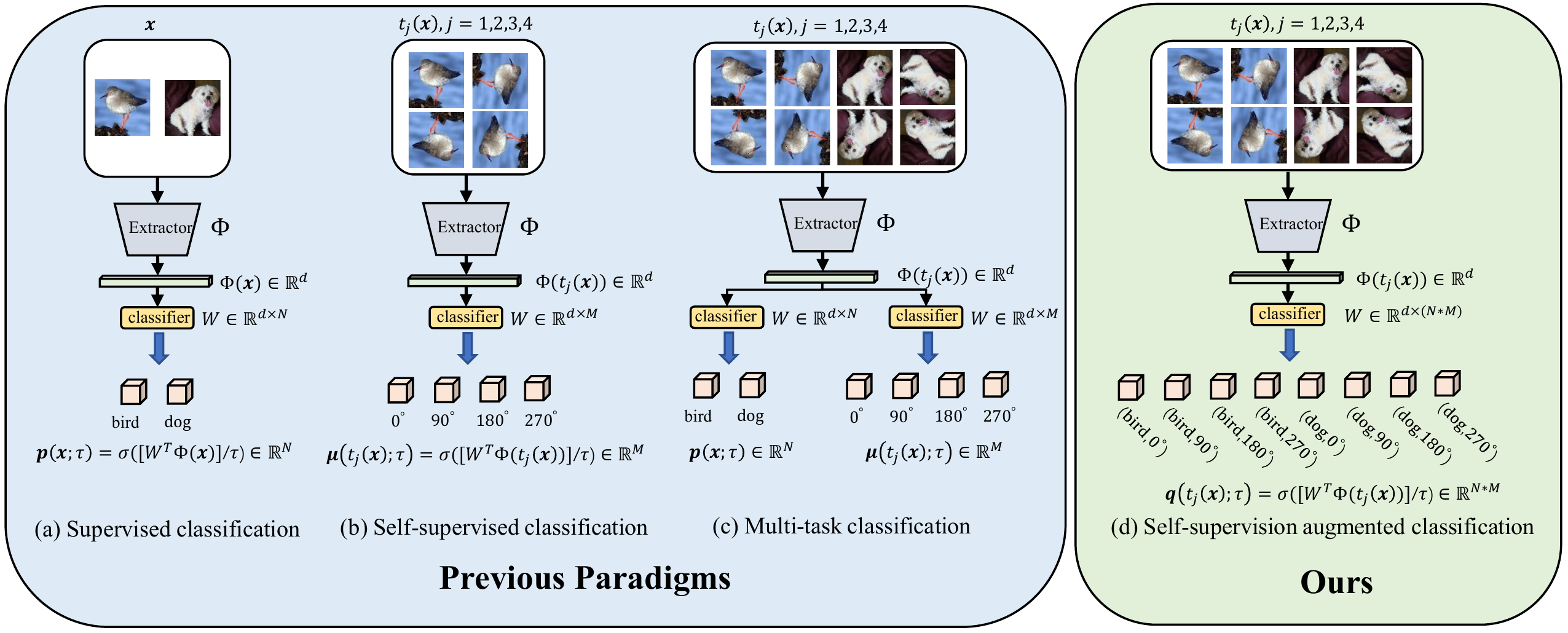}
	\caption{An overview of (a) $N$-way supervised classification, (b) $M$-way self-supervised classification, (c) multi-task learning by performing $N$-way supervised and  $M$-way self-supervised classification respectively, (d) ($N*M$)-way self-supervision augmented classification. Here, we employ ($N$=2)-way supervised \{bird, dog\} and ($M$=4)-way self-supervised $\{0^{\circ},90^{\circ},180^{\circ},270^{\circ}\}$ classification for illustration. }
	\label{classification}
\end{figure*}
\subsection{Self-Supervision Augmented Distribution}
\label{four}
In this section, we describe the details of our proposed self-supervision augmented distribution. Before this, we first review the conventional \emph{supervised classification task}, \emph{self-supervised classification task} and \emph{multi-task learning}. The overview is illustrated in Fig.~\ref{classification}.

  \subsubsection{$N$-way supervised classification task} We consider a training set $\mathcal{X}$ with fully-annotated label space $\mathcal{Y}=\{1,2,\cdots,N\}$, where $\bm{x}\in \mathcal{X}$ is an input sample with the label $y\in \mathcal{Y}$, $N$ is the number of classes. We aim to train an ordinary CNN $f(\cdot)$ to map the input sample $\bm{x}$ to a predictive class probability distribution $\bm{p}\in \mathbb{R}^{N}$, \emph{i.e.} $f:\bm{x}\rightarrow \bm{p}$. $f(\cdot)$ could be decomposed of a feature extractor $\Phi(\cdot)$ and a linear classifier $W$. Given an input $\bm{x}$, $\Phi(\cdot)$ encodes it into a feature embedding vector $\Phi(\bm{x}) \in \mathbb{R}^{d}$, where $d$ is the embedding size. Afterwards, a linear matrix $W\in \mathbb{R}^{d\times N}$ transforms the feature vector $\Phi(\bm{x}) \in \mathbb{R}^{d}$ into a class logits distribution $f(\bm{x})=W^{T}\Phi(\bm{x}) \in \mathbb{R}^{N}$. Following the common protocol, we scale the logits distribution $f(\bm{x})$ with a temperature hyper-parameter $\tau$ and then normalize it by a \emph{softmax} function $\sigma$ to derive the final class probability distribution:
  \begin{equation} 
  \bm{p}(\bm{x};\tau)=\sigma(f(\bm{x})/\tau)=\sigma([W^{T}\Phi(\bm{x})]/\tau)\in \mathbb{R}^{N}.
  \end{equation}

\subsubsection{$M$-way self-supervised classification task}  In principle, the self-supervised task can be regarded as a classification problem to deal with human pre-defined transformations. This framework has been successfully applied to self-supervised feature representation learning, \emph{i.e.} random rotations~\cite{gidaris2018unsupervised} and colorful image colorization~\cite{zhang2016colorful}. Formally, we define $M$ various image transformations $\{t_{j}\}_{j=1}^{M}$ over an input image $\bm{x}$ with the label space $\mathcal{M}=\{1,\cdots,M\}$, where $t_{1}$ denotes the identity transformation, \emph{i.e.} $t_{1}(\bm{x})=\bm{x}$. We encourage the network to correctly recognize which transformation $t_{j}$ is applied to the transformed image. Given an input sample $t_{j}(\bm{x})$, the final output is a $M$-way class probability distribution denoted as:
\begin{equation}
	\bm{\mu}(t_{j}(\bm{x}; \tau))=\sigma(f(t_{j}(\bm{x})/\tau)=\sigma([W^{T}\Phi(t_{j}(\bm{x}))]/\tau)\in \mathbb{R}^{M}.
\end{equation}
Here, $W\in \mathbb{R}^{d\times M}$ is the linear classifier for $M$-way self-supervised classification.

\subsubsection{Multi-task learning} To learn feature representations benefiting from both supervised and self-supervised tasks, a natural idea is to employ multi-task learning. In practice, we can use a shared feature extractor and separated classifier heads to perform $N$-way supervised and  $M$-way self-supervised classification, respectively.

\subsubsection{($N*M$)-way self-supervision augmented classification} 
Although we aim to solve a supervised classification task, we can further introduce an additional self-supervised task to enrich the class space. We combine supervised and self-supervised tasks into a unified task to guide the network learning a joint distribution. The label space of this unified task can be expressed as $\mathcal{N}\otimes \mathcal{M}$, where $\otimes$ is a Cartesian product. $\left | \mathcal{N}\otimes \mathcal{M} \right |=N*M$, where $\left | \cdot \right |$ is the cardinality of the label collection and * denotes element multiplication. Intuitively, a more competitive joint task would require the network to generate more informative and meaningful feature representations compared with a single task. Given a transformed input image $t_{j}(\bm{x}),j\in \mathcal{M}$, Similar to the practice of the conventional classification problem, we can derive the final self-supervision augmented distribution as:
\begin{equation}
\bm{q}(t_{j}(\bm{x});\tau)=\sigma(f(t_{j}(\bm{x}))/\tau)=\sigma([W^{T}\Phi(t_{j}(\bm{x}))]/\tau)\in \mathbb{R}^{N*M}
\end{equation}
 over the joint label space $\mathcal{N}\otimes \mathcal{M}$. Here, $W\in \mathbb{R}^{d\times (N*M)}$ is the linear classifier for $N*M$-way self-supervision augmented classification. In this paper, we find our proposed self-supervision augmented distribution is the best type of probabilistic knowledge to learn or conduct KD-based feature representation learning.


\begin{table*}
	\centering
	\caption{Architectural details of a ResNet-56~\cite{he2016deep} as the backbone $f(\cdot)$ with its three auxiliary branches $\{c_{l}(\cdot)\}_{l=1}^{3}$. We force the backbone $f(\cdot)$ to output the primary $N$-way supervised class probability distribution and three auxiliary branches $c_{1}(\cdot)\sim c_{3}(\cdot)$ to output $(N*M)$-way self-supervision augmented distributions.}
	\resizebox{0.8\linewidth}{!}{
		\begin{tabular}{c|c|c|c|c|c}
			\hline
			Layer Name& Spatial Size& $f(\cdot)$& $c_{1}(\cdot)$& $c_{2}(\cdot)$& $c_{3}(\cdot)$\\ 	\hline
			Stem&  32$\times$32 &  $3\times3, 16$ &-&-&-\\  \hline
			Stage1 &  32$\times$32        & 
			$\begin{bmatrix}
			3\times 3, 16\\ 
			3\times 3, 16
			\end{bmatrix}\times 9 $  &  -&  - &  -  \\  \hline
			Stage2 &  16$\times$16        & 
			$\begin{bmatrix}
			3\times 3, 32\\ 
			3\times 3, 32
			\end{bmatrix}\times 9 $  &  $\begin{bmatrix}
			3\times 3, 32\\ 
			3\times 3, 32
			\end{bmatrix}\times 9 $ & - & -  \\  \hline
			Stage3 &  8$\times$8        & 
			$\begin{bmatrix}
			3\times 3, 64\\ 
			3\times 3, 64
			\end{bmatrix}\times 9 $  &  $\begin{bmatrix}
			3\times 3, 64\\ 
			3\times 3, 64
			\end{bmatrix}\times 9 $&  $\begin{bmatrix}
			3\times 3, 64\\ 
			3\times 3, 64
			\end{bmatrix}\times 9 $& -  \\  \hline
			-&  8$\times$8          & - &  -&  - &  $\begin{bmatrix}
			3\times 3, 64\\ 
			3\times 3, 64
			\end{bmatrix}\times 9 $  \\  \hline
			Pooling &  1$\times$1  &  \multicolumn{4}{c}{Global Average Pooling}  \\  \hline
			Linear Classifier &  1$\times$1  &  $N$-D FC&  $(N*M)$-D FC &  $(N*M)$-D FC&  $(N*M)$-D FC  \\  \hline
	\end{tabular}}
	
	\label{resnet56}
\end{table*}

\subsection{Auxiliary Architecture Design}
It has been widely known that a general CNN gradually downsamples the feature maps along the spatial dimension to refine meaningful semantics. Feature maps with various spatial resolutions produced from different convolutional stages often encode various patterns of representational information. Higher-resolution feature maps from low-level stages often present more fine-grained object details and high-frequency information favourable for small object detection and fine-grained image classification. In contrast, lower-resolution ones from high-level stages extracted by a larger receptive field often contain richer global structures and semantic information. They may lead to better recognition performance by reducing the redundant details. Leveraging multi-scale feature information is demonstrated as an effective practice in previous works~\cite{lin2017feature,sun2019deep}. To take full advantage of hierarchical feature maps encapsulated in a single network, we append several intermediate auxiliary branches into hidden layers. These branches are responsible for generating and distilling multi-level self-supervision augmented distributions from multi-scale feature maps.

Modern CNNs typically utilize stage-wise convolutional blocks to gradually extract coarser features as the depth of the network increases. For example, the popular ResNet-50 for ImageNet classification contains consecutive four stages, and extracted feature maps produced from various stages have different granularities and patterns. Formally, we describe how to insert auxiliary branches into a conventional CNN in a heuristic principle. Given a CNN $f(\cdot)$ that consists of $L$ stages, we choose to append an auxiliary branch after each stage, thus resulting in $L$ branches $\{c_{l}(\cdot)\}_{l=1}^{L}$, where $c_{l}(\cdot)$ denotes the auxiliary after the $l$-th stage. Previous deeply-supervised nets~\cite{lee2015deeply} inspire this practice. 

However, appending simply linear auxiliary classifiers without extra feature extraction modules may not be helpful to explore meaningful information, as empirically verified by~\cite{huang2017multi}. Therefore, we try to design more complex auxiliary branches to provide informative task-aware knowledge as intended. Specifically, each auxiliary branch consists of a feature extraction module, a global average pooling layer and a linear classifier with a desirable dimension for the used auxiliary task, \emph{e.g.} $N*M$ for our self-supervision augmented task. The feature extraction module is composed of several building blocks as same as the original backbone network, \emph{i.e.} residual block in ResNets~\cite{he2016deep}. As empirically shown in~\cite{huang2017multi}, early layers may lack coarse features that encapsulate high-level semantic information, which is crucial for image-level recognition. To ensure the appropriate fine-to-coarse feature transformation, we make the path from an input to the end of an auxiliary branch with the same downsampling number as the backbone network. In Table~\ref{resnet56}, we systematically show the overall architectural details of a ResNet-56~\cite{he2016deep} as the backbone $f(\cdot)$ with its three auxiliary branches $\{c_{l}(\cdot)\}_{l=1}^{3}$ as an example for instantiating our design principle. Fig.~\ref{SMG} further illustrates the overall architecture schematically.

Formally, we describe the inference graph of the backbone network $f(\cdot)$ and its $L$ attached auxiliary branches $\{c_{l}(\cdot)\}_{l=1}^{L}$. Given a transformed input sample $t_{j}(\bm{x})$, where $j\in \mathcal{M}$ and $t_{1}(\bm{x})=\bm{x}$, the network $f(t_{j}(\bm{x}))$ outputs $L$ intermediate feature-maps $\{\bm{F}_{l,j}\}_{l=1}^{L}$ from $L$ convolutional stages respectively and a final primary class probability distribution $\bm{p}(t_{j}(\bm{x});\tau)=\sigma(f(t_{j}(\bm{x}))/\tau)\in \mathbb{R}^{N}$ over the original class space $\mathcal{N}$. We further feed $\{\bm{F}_{l,j}\}_{l=1}^{L}$ to the corresponding $L$ auxiliary branches $\{c_{l}(\cdot)\}_{l=1}^{L}$ to learn desirable auxiliary tasks, respectively. Given the $\bm{F}_{l,j}$, the $l$-th branch $c_{l}(\cdot)$ can output a type of distribution from a specific auxiliary task, which is formulated as follows:
\begin{itemize}
	\item a supervised class probability distribution
	\begin{equation}
	\bm{p}_{l}(\bm{x};\tau)=\sigma(c_{l}(\bm{F}_{l,j})/\tau)\in \mathbb{R}^{N}, s.t.\ j=1, 
	\end{equation}
	over the class space $\mathcal{N}$;
	
	\item a self-supervised class probability distribution
	\begin{equation}
	\bm{\mu}_{l}(t_{j}(\bm{x});\tau)=\sigma(c_{l}(\bm{F}_{l,j})/\tau)\in \mathbb{R}^{M}, s.t.\ 1\leqslant j\leqslant M,
	\end{equation}
	over the class space $ \mathcal{M}$;
	
	\item a self-supervision augmented distribution
	\begin{equation}
		\bm{q}_{l}(t_{j}(\bm{x});\tau)=\sigma(c_{l}(\bm{F}_{l,j})/\tau)\in \mathbb{R}^{N*M}, s.t.\ 1\leqslant j\leqslant M,
	\end{equation}
	over the joint class space $\mathcal{N}\otimes \mathcal{M}$.
\end{itemize}
As referred above, we can change the dimension of the linear classifier in the $l$-th auxiliary branch $c_{l}$ to deal with various classification tasks.

\subsection{Training a Single Network with Auxiliary Branches}
\label{C}
\begin{figure*}
	\centering 
	\begin{subfigure}[t]{0.25\textwidth}
		\centering
		\includegraphics[width=\textwidth]{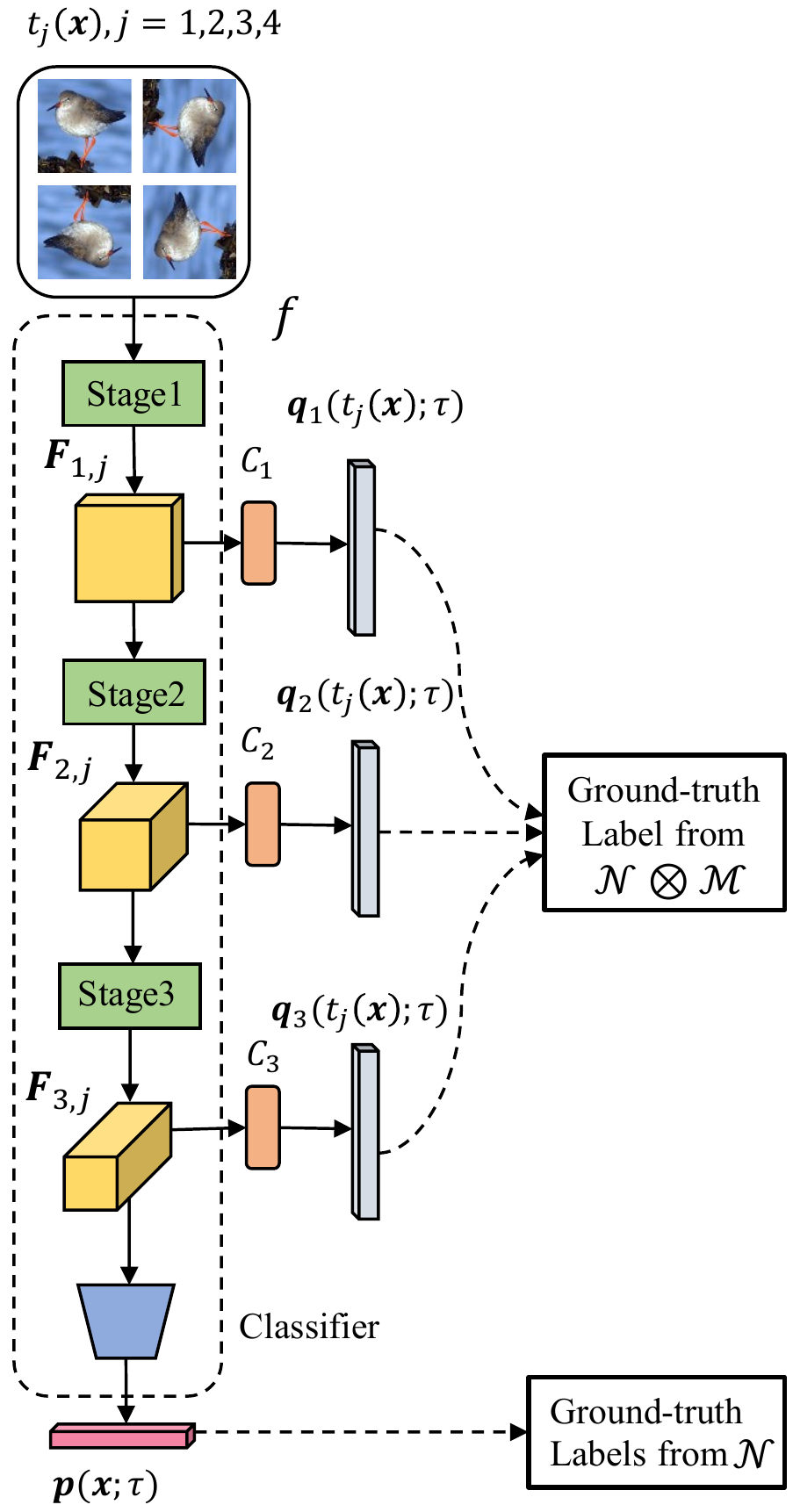}
		\caption{Training a single network.}
		\label{SMG}
	\end{subfigure}
	\begin{subfigure}[t]{0.33\textwidth}
		\centering
		\includegraphics[width=\textwidth]{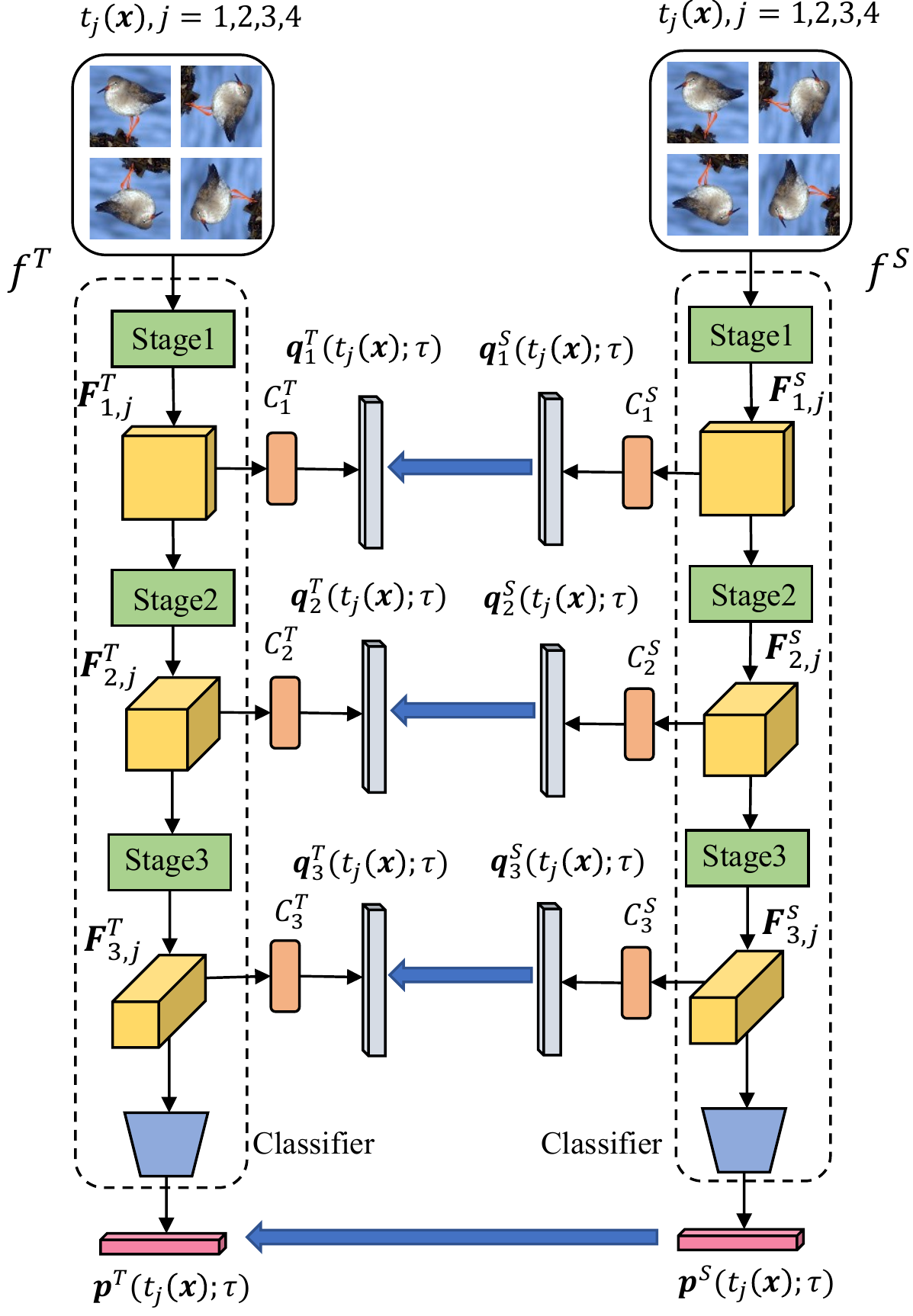}
		\caption{Teacher-student based offline HSSAKD.}
		\label{Update}
	\end{subfigure}
	\begin{subfigure}[t]{0.336\textwidth}
		\centering
		\includegraphics[width=\textwidth]{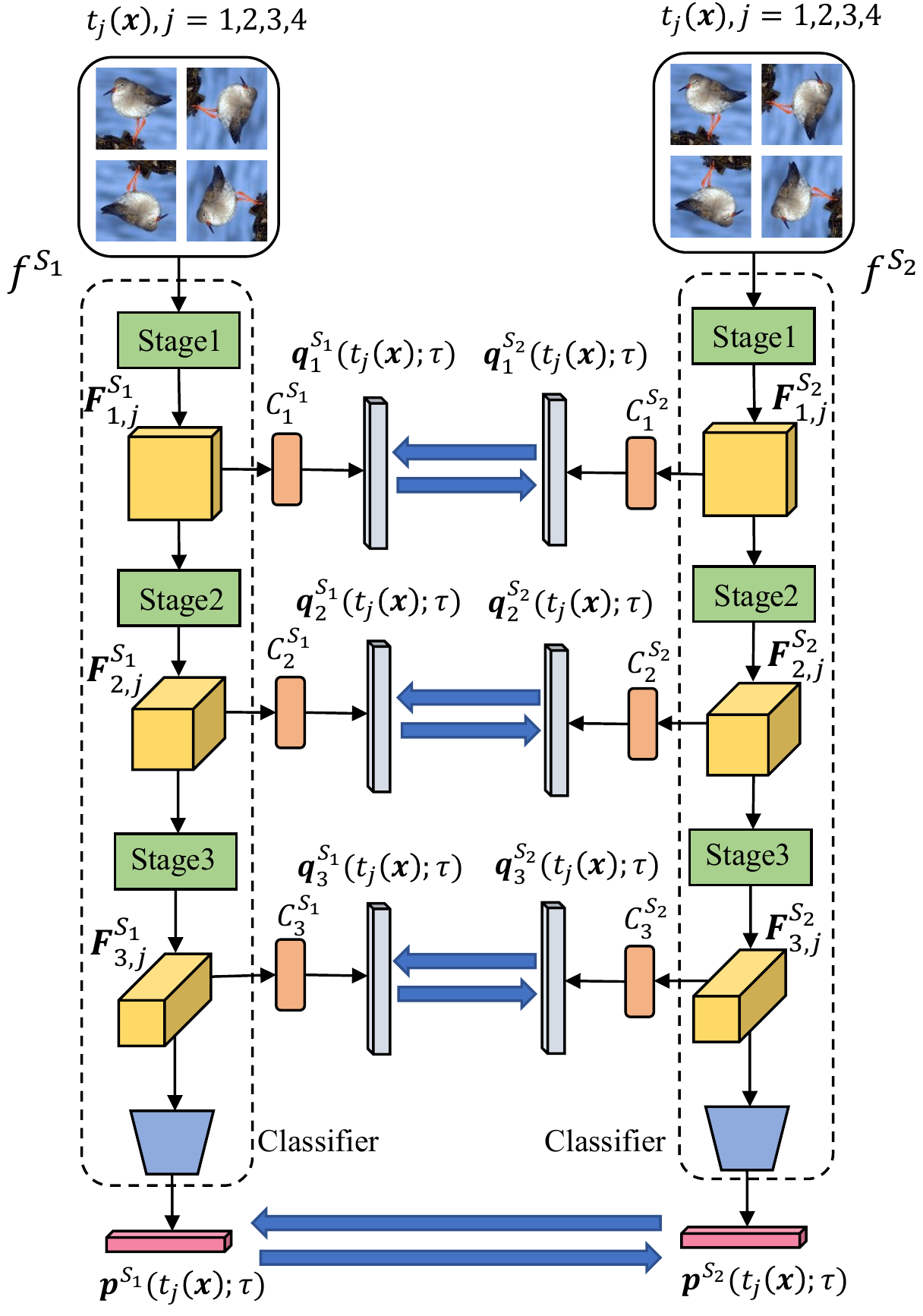}
		\caption{Multi-student based online HSSAKD.}
		\label{Forget}
	\end{subfigure}
	\caption{\textbf{Overview of using our proposed self-supervision augmented distributions under three scenarios}. The blue arrow denotes the mimicry direction between two probability distributions by KL-divergence. }  
	\label{Overview_methods} 
\end{figure*}

Given a single network backbone $f(\cdot)$ and its attached several auxiliary branches $\{c_{l}(\cdot)\}_{l=1}^{L}$, we train them jointly in an end-to-end manner, as shown in Fig.~\ref{SMG}. Given an input sample $\bm{x}$, we guide the $f(\cdot)$ to fit the ground-truth label $y\in \mathcal{N}$ using cross-entropy as the primary task loss:
\begin{equation}
\mathcal{L}_{task}=\mathcal{L}_{ce}(\bm{p}(\bm{x};\tau),y).
\label{ce_cpd}
\end{equation}
Where $\mathcal{L}_{ce}$ denotes the cross-entropy loss and $\bm{p}(\bm{x};\tau)$ is the supervised class probability distribution from $f$, \emph{i.e.}, $\bm{p}(\bm{x};\tau)=\sigma(f(\bm{x})/\tau)\in \mathbb{R}^{N}$. This loss aims to make $f$ fit the normal data for learning general classification capability and task-specific features.

For auxiliary branches, we can guide them to learn additional tasks, as referred in Section~\ref{four}. As for all $L$ auxiliary branches, the losses of four tasks are respectively formulated as follow:
\begin{itemize}
	\item \emph{$N$-way supervised classification task}. This task is identical to the primary task. The loss for training $L$ auxiliary branches is formulated as:
	\begin{equation}
	\mathcal{L}_{aux\_SCPD}=\sum_{l=1}^{L}\mathcal{L}_{ce}(\bm{p}_{l}(\bm{x};\tau),y).
	\label{aux_SCPD}
	\end{equation}
	Here, $\bm{p}_{l}(\bm{x};\tau)\in \mathbb{R}^{N}$ is the predictive \emph{Supervised Class Probability Distributions} (SCPD) from the $l$-th auxiliary branch.
	\item \emph{$M$-way self-supervised classification task}. We can apply $M$ transformations $\{t_{j}(\cdot)\}_{j=1}^{M}$ to the input sample $\bm{x}$ to construct $M$ synthetic samples $\{t_{j}(\bm{x})\}_{j=1}^{M}$. We expect that the classifier can correctly recognize which transformation is applied to the synthetic sample. The loss for training $L$ auxiliary branches is formulated as:
	\begin{equation}
	\mathcal{L}_{aux\_SSCPD}=\frac{1}{M}\sum_{j=1}^{M}\sum_{l=1}^{L}\mathcal{L}_{ce}(\bm{\mu}_{l}(t_{j}(\bm{x});\tau),j).
	\label{aux_SSCPD}
	\end{equation}
	Here, $\bm{\mu}_{l}(t_{j}(\bm{x});\tau)\in \mathbb{R}^{M}$ is the predictive \emph{Self-Supervised Class Probability Distributions} (SSCPD) from the $l$-th auxiliary classifier. $j\in \mathcal{M}$ is the ground-truth label for the transformed sample $t_{j}(\bm{x})$. Inspired by~\cite{Spyros2018Unsupervised}, this loss would enable the network to learn self-supervised feature representations.
	
	\item \emph{Multi-task classification task}. We can use two separated classification heads in each auxiliary branch to perform $N$-way supervised task and $M$-way self-supervised task, respectively. The loss for training $L$ auxiliary branches is formulated as:
	\begin{align}
	\label{aux_multi_task}
	\mathcal{L}_{aux\_multi\_task}&=\mathcal{L}_{aux\_SCPD}+\mathcal{L}_{aux\_SSCPD} \\
	&=\sum_{l=1}^{L}\mathcal{L}_{ce}(\bm{p}_{l}(\bm{x};\tau),y) \notag \\
	&+\frac{1}{M}\sum_{j=1}^{M}\sum_{l=1}^{L}\mathcal{L}_{ce}(\bm{\mu}_{l}(t_{j}(\bm{x});\tau),j). \notag
	\end{align}
	
	\item \emph{$(N*M)$-way self-supervision augmented task}. We guide all auxiliary branches $\{c_{l}(\cdot)\}_{l=1}^{L}$ to learn self-supervision augmented tasks using cross-entropy to fit the joint ground-truth label $(y,j)\in \mathcal{N}\otimes \mathcal{M}$. The loss for training $L$ auxiliary branches is formulated as:
	
	\begin{equation}
	\mathcal{L}_{aux\_SSAD}=\frac{1}{M}\sum_{j=1}^{M}\sum_{l=1}^{L}\mathcal{L}_{ce}(\bm{q}_{l}(t_{j}(\bm{x});\tau),(y,j)).
	\label{ce_ssad}
	\end{equation}
	Here, $\bm{\bm{q}}_{l}(t_{j}(\bm{x});\tau)\in \mathbb{R}^{N*M}$ is the predictive \emph{Self-Supervised Augmented Distributions} (SSAD) from the $l$-th auxiliary classifier. This loss forces auxiliary classifiers to solve the unified recognition of the categories of objects and transformations. It can implicitly enable the network to learn more informative and meaningful feature representations by dealing with a more competitive classification task.
\end{itemize}


We jointly train the backbone $f(\cdot)$ and $L$ auxiliary branches $\{c_{l}(\cdot)\}_{l=1}^{L}$ by summarizing the primary task loss $\mathcal{L}_{task}$ and an auxiliary loss  $\mathcal{L}_{aux}$ as:
\begin{equation}
\mathcal{L}^{single}=\mathbb{E}_{\bm{x}\in \mathcal{X}}(\mathcal{L}_{task}+\mathcal{L}_{aux}).
\label{ce_single}
\end{equation}
Here, $\mathcal{X}$ is the training set and $\mathcal{L}_{aux}$ can be any loss from $\{\mathcal{L}_{aux\_SCPD},\mathcal{L}_{aux\_SSCPD},\mathcal{L}_{aux\_multi\_task},\mathcal{L}_{aux\_SSAD}\}$. During the test stage, we only retain the backbone network $f(\cdot)$ and discard $L$ auxiliary branches $\{c_{l}(\cdot)\}_{l=1}^{L}$, leading to no extra inference costs compared with the original baseline network.

As validated in Section~\ref{training_single}, our proposed auxiliary loss $\mathcal{L}_{aux\_SSAD}$ can significantly improve the primarily supervised classification accuracy of the backbone network and surpass other alternative auxiliary losses by large margins. The results further inspire us to introduce $(N*M)$-way self-supervision augmented distribution as promising knowledge to the field of KD. 

\subsection{Teacher-Student Based Offline HSSAKD}
The conventional KD aims to train a student from scratch by distilling meaningful knowledge from a pre-trained teacher network. Similar to the consistent protocol, we formulate our proposed offline HSSAKD by introducing a teacher network  $f^{T}(\cdot)\cup  \{c_{l}^{T}(\cdot)\}_{l=1}^{L}$ and a student network $f^{S}(\cdot)\cup  \{c_{l}^{S}(\cdot)\}_{l=1}^{L}$. We conduct a two-stage pipeline: (1) training the teacher network at first; (2) training the student network under the supervision of the pre-trained frozen teacher network. Fig.~\ref{Update} illustrates the overview schematically.

\subsubsection{Training the Teacher Network} 
We denote the $\bm{p}^{T}(\bm{x};\tau)\in \mathbb{R}^{N}$ as the predictive class probaility distribution generated from $f^{T}(\cdot)$ and self-supervised augmented distributions $\{\bm{q}_{l}^{T}(t_{j}(\bm{x});\tau)\}_{l=1}^{L}$ generated from $\{c_{l}^{T}(\cdot)\}_{l=1}^{L}$ respectively. We train the teacher network using the primary task loss and an auxiliary loss referred in Section~\ref{C}, which is formulated as $\mathcal{L}^{T}$:
\begin{align}
\mathcal{L}^{T}=\mathbb{E}_{\bm{x}\in \mathcal{X}}&[\mathcal{L}_{task}^{T}+\mathcal{L}^{T}_{aux\_SSAD}] \\
=\mathbb{E}_{\bm{x}\in \mathcal{X}}&[\mathcal{L}_{ce}(\bm{p}^{T}(\bm{x};\tau),y) \notag\\
&+\frac{1}{M}\sum_{j=1}^{M}\sum_{l=1}^{L}\mathcal{L}_{ce}(\bm{q}^{T}_{l}(t_{j}(\bm{x});\tau),(y,j))]. \notag
\label{L_T}
\end{align}
\subsubsection{Training the Student Network} 
We conduct an end-to-end training process under the supervision of the teacher network. The overall loss includes a primary task loss from pre-defined ground-truth labels and mimicry losses from the pre-trained teacher network. 

\paragraph{Task loss} Given an input sample $\bm{x}$, we train the student backbone $f^{S}(\cdot)$ to fit the ground-truth label $y\in \mathcal{N}$ using the cross-entropy loss:
\begin{equation}
\mathcal{L}_{task}^{S}=\mathcal{L}_{ce}(\bm{p}^{S}(\bm{x};\tau),y).
\label{sce_cpd}
\end{equation}
Where $\bm{p}^{S}(\bm{x};\tau)$ is the predictive class probability distribution generated from the $f^{S}(\cdot)$.

\paragraph{Mimicry loss from auxiliary self-supervision augmented distributions} We consider forcing hierarchical self-supervision augmented distributions $\{\bm{q}^{S}_{l}(t_{j}(\bm{x});\tau)\}_{l=1}^{L}$ from $L$ auxiliary branches of the student network to mimic corresponding $\{\bm{q}^{T}_{l}(t_{j}(\bm{x});\tau)\}_{l=1}^{L}$ generated from $L$ auxiliary branches of the teacher network, respectively. This distillation process performs in a one-to-one manner by KL-divergence function $D_{\rm{KL}}$. The loss is formulated as Eq.~(\ref{mini_ss}), where the weight coefficient $\tau^{2}$ is used to retain the gradient contributions unchanged as the temperature $\tau$ varies according to the KD recommendations~\cite{hinton2015distilling}.
\begin{equation}
\mathcal{L}_{KL\_q}^{S}=\frac{1}{M}\sum_{j=1}^{M}\sum_{l=1}^{L}\tau^{2}D_{\rm{KL}}(\bm{q}^{T}_{l}(t_{j}(\bm{x});\tau)\parallel \bm{q}^{S}_{l}(t_{j}(\bm{x});\tau)).
\label{mini_ss}
\end{equation}
Benefiting from Eq.~(\ref{mini_ss}), one can expect that the student network gains comprehensive guidances of unified self-supervised knowledge and the original supervised knowledge. The informative self-supervision augmented knowledge is derived from multi-scale intermediate feature maps encapsulated in hidden layers of the high-capacity teacher network. It leads to better representation learning for various-stage features of the student network.

\paragraph{Mimicry loss from the primary supervised class probability distribution} Similar to the conventional KD, we also transfer the supervised class probability distributions generated from the final layer between teacher $f^{T}(\cdot)$ and student $f^{S}(\cdot)$. Specifically, we transfer the class posterior derived from both the normal and transformed data denoted as $\{t_{j}(\bm{x})\}_{j=1}^{M}$, where $t_{1}(\bm{x})=\bm{x}$. This loss is formulated as Eq.~(\ref{mini_s}). 
\begin{equation}
\mathcal{L}_{KL\_p}^{S}=\frac{1}{M}\sum_{j=1}^{M}\tau^{2}D_{\rm{KL}}(\bm{p}^{T}(t_{j}(\bm{x});\tau)\parallel \bm{p}^{S}(t_{j}(\bm{x});\tau)).
\label{mini_s}
\end{equation}
It can be observed that we do not explicitly force the student backbone $f^{S}(\cdot)$ to fit the transformed data in task loss for preserving the normal classification capability. But we argue that mimicking the side product of $\{\bm{p}^{T}(t_{j}(\bm{x})\}_{j=2}^{M}$  from extra transformed data inferred from the teacher network is also beneficial for the feature representation learning of the student network. As validated in Section~\ref{kd_loss}, Eq.~(\ref{mini_s}) can result in better performance than only distilling $\bm{p}^{T}(t_{1}(\bm{x});\tau)$ as same as the original KD.

\paragraph{Overall loss} We summarize the task loss and mimicry loss as the overall loss $\mathcal{L}^{S}_{KD}$ for training the student network.
\begin{equation}
\mathcal{L}^{S}_{KD}=\mathbb{E}_{\bm{x}\in \mathcal{X}}(\mathcal{L}_{task}^{S}+\mathcal{L}_{KL\_q}^{S}+\mathcal{L}_{KL\_p}^{S}).
\label{overall}
\end{equation}
Following the broad practice, we set the hyper-parameter $\tau=1$ in cross-entropy based task loss and $\tau=3$ in KL-based mimicry loss, and we do not introduce other coefficients about loss weights. Empirically, we found this practice sufficient to work good enough.

\subsubsection{Testing the Student Network}
During the test stage, we only retain the student backbone network $f^{S}(\cdot)$ and discard $L$ auxiliary branches $\{c_{l}^{S}(\cdot)\}_{l=1}^{L}$ as well as the teacher network  $f^{T}(\cdot)\cup  \{c_{l}^{T}(\cdot)\}_{l=1}^{L}$, leading to no extra inference costs compared with the original baseline network.

\subsection{Multi-Student Based Online HSSAKD}
Unlike the teacher-student based KD, online KD does not need a pre-trained teacher model as a prerequisite. Instead, it aims to train a cohort of student models collaboratively from scratch. During the training stage, multiple students teach each other and perform knowledge transfer in an online manner. Therefore, online KD is an end-to-end training process rather than a two-stage pipeline in the conventional KD. We demonstrate that our offline HSSAKD can be extended to online HSSAKD. Fig.~\ref{Forget} illustrates the overview schematically.

In this section, we formulate the training graph as $K(K\geqslant 2)$ peer networks denoted as $\{f^{S_{k}}(\cdot)\cup  \{c_{l}^{S_{k}}(\cdot)\}_{l=1}^{L}\}_{k=1}^{K}$, where the $k$-th network consists of a backbone $f^{S_{k}}(\cdot)$ and $L$ auxiliary branches $\{c_{l}^{S_{k}}(\cdot)\}_{l=1}^{L}$. For each network, the training loss includes a primary task loss, an auxiliary task loss and mimicry losses. 
\subsubsection{Primary and auxiliary task losses} For the $k$-th network, we denote the $\bm{p}^{S_{k}}(\bm{x};\tau)\in \mathbb{R}^{N}$ as the primary supervised class probaility distribution generated from $f^{S_{k}}(\cdot)$ and $L$ auxiliary self-supervision augmented distributions $\{\bm{q}_{l}^{S_{k}}(t_{j}(\bm{x});\tau)\}_{l=1}^{L}$ generated from $\{c_{l}^{S_{k}}(\cdot)\}_{l=1}^{L}$ respectively. For each network, we apply the same task loss and auxiliary loss referred in Section~\ref{C}. The total loss for training $M$ networks are reformulated as Eq.~(\ref{online_task}) and Eq.~(\ref{online_aux}), respectively:
\begin{equation}
\mathcal{L}^{S_{1}\sim S_{K}}_{task}=\sum_{k=1}^{K}\mathcal{L}_{ce}(\bm{p}^{S_{k}}(\bm{x};\tau),y),
\label{online_task}
\end{equation}

\begin{equation}
\mathcal{L}^{S_{1}\sim S_{K}}_{aux\_SSAD}=\frac{1}{M}\sum_{k=1}^{K}\sum_{j=1}^{M}\sum_{l=1}^{L}\mathcal{L}_{ce}(\bm{q}^{S_{k}}_{l}(t_{j}(\bm{x});\tau),(y,j)).
\label{online_aux}
\end{equation}

\subsubsection{Mutual mimicry loss between peer auxiliary self-supervision augmented distributions} DML~\cite{zhang2018deep} indicates that a cohort of student networks can benefit from transferring class probability distributions mutually in an online manner. Such a peer-teaching method makes a network learn better than training alone using the conventional supervised learning scenario. Inspired by this basic idea, we consider mutually mimicking our proposed self-supervision augmented distributions for online KD. Unlike the conventional DML that performs online KD only in the final layers, our HSSAKD considers more comprehensive knowledge derived from hierarchical feature maps encoded in hidden layers.

Taking two student networks $f^{S_{a}}$ and $f^{S_{b}}$ ($a\neq b$, $a,b\in [1,K]$) for example, we guide hierarchical self-supervision augmented distributions $\{\bm{q}^{S_{a}}_{l}(t_{j}(\bm{x});\tau)\}_{l=1}^{L}$ from $L$ auxiliary branches of the $S_{a}$ network to mimic corresponding $\{\bm{q}^{S_{b}}_{l}(t_{j}(\bm{x});\tau)\}_{l=1}^{L}$ generated from $L$ auxiliary branches of the $S_{b}$ network respectively, as shown in Eq.~(\ref{mini_ss_online1}):

\begin{equation}
\mathcal{L}_{KL\_q}^{S_{a}\rightarrow  S_{b}}=\frac{1}{M}\sum_{j=1}^{M}\sum_{l=1}^{L}\tau^{2}D_{\rm{KL}}(\bm{q}^{S_{b}}_{l}(t_{j}(\bm{x});\tau)\parallel \bm{q}^{S_{a}}_{l}(t_{j}(\bm{x});\tau)),
\label{mini_ss_online1}
\end{equation}
 and vice versa in Eq.~(\ref{mini_ss_online2}):
 \begin{equation}
 \mathcal{L}_{KL\_q}^{S_{b}\rightarrow  S_{a}}=\frac{1}{M}\sum_{j=1}^{M}\sum_{l=1}^{L}\tau^{2}D_{\rm{KL}}(\bm{q}^{S_{a}}_{l}(t_{j}(\bm{x});\tau)\parallel \bm{q}^{S_{b}}_{l}(t_{j}(\bm{x});\tau))
 \label{mini_ss_online2}
 \end{equation}
This bidirectional distillation process performs in a one-to-one manner by KL-divergence function $D_{\rm{KL}}$.

As suggested by \emph{Miyato et al.}~\cite{miyato2018virtual}, the gradient is not propagated through soft labels to avoid the model collapse problem. Therefore, we stop the gradient propagation of $\partial \mathcal{L}_{KL\_q}^{S_{a}\rightarrow  S_{b}}/ \partial \bm{q}^{S_{b}}_{l}(t_{j}(\bm{x});\tau)$ in Eq.~(\ref{mini_ss_online1}) and $\partial \mathcal{L}_{KL\_q}^{S_{b}\rightarrow  S_{a}}/ \partial \bm{q}^{S_{a}}_{l}(t_{j}(\bm{x});\tau)$ in Eq.~(\ref{mini_ss_online2}).

When $K\geqslant 2$, we perform mutual mimicry every two of $K$ student networks to learn more diverse knowledge captured by other networks, leading to the total loss as:
\begin{equation}
\mathcal{L}_{KL\_q}^{S_{1}\sim  S_{K}}=\sum_{1\leqslant a,b \leqslant K}(\mathcal{L}_{KL\_q}^{S_{a}\rightarrow  S_{b}}+\mathcal{L}_{KL\_q}^{S_{b}\rightarrow  S_{a}}).
\end{equation}

\subsubsection{Mutual mimicry loss between primary supervised class probability distributions} We also conduct mutual mimicry of supervised class probability distributions generated from the final layers among multiple student networks. These distributions are also derived from both the normal and transformed data denoted as $\{t_{j}(\bm{x})\}_{j=1}^{M}$. For two peer networks $f^{S_{a}}$ and $f^{S_{b}}$ ($a\neq b$, $a,b\in [1,K]$), we guide $\{\bm{p}^{S_{a}}(t_{j}(\bm{x};\tau)\}_{j=1}^{M}$ generated from $f^{S_{a}}$ to mimic $\{\bm{p}^{S_{b}}(t_{j}(\bm{x};\tau)\}_{j=1}^{M}$ generated from $f^{S_{b}}$, as shown in Eq.~(\ref{mini_sup_online1}):
\begin{equation}
\mathcal{L}_{KL\_p}^{S_{a}\rightarrow  S_{b}}=\frac{1}{M}\sum_{j=1}^{M}\tau^{2}D_{\rm{KL}}(\bm{p}^{S_{b}}(t_{j}(\bm{x});\tau)\parallel \bm{p}^{S_{a}}(t_{j}(\bm{x});\tau)),
\label{mini_sup_online1}
\end{equation}
 and vice versa in Eq.~(\ref{mini_sup_online2}):
\begin{equation}
\mathcal{L}_{KL\_p}^{S_{b}\rightarrow  S_{a}}=\frac{1}{M}\sum_{j=1}^{M}\tau^{2}D_{\rm{KL}}(\bm{p}^{S_{a}}(t_{j}(\bm{x});\tau)\parallel \bm{p}^{S_{b}}(t_{j}(\bm{x});\tau)).
\label{mini_sup_online2}
\end{equation}

Here, we stop the gradient propagation of $\partial \mathcal{L}_{KL\_p}^{S_{a}\rightarrow  S_{b}}/ \partial \bm{p}^{S_{b}}(t_{j}(\bm{x});\tau)$ in Eq.~(\ref{mini_sup_online1}) and $\partial \mathcal{L}_{KL\_q}^{S_{b}\rightarrow  S_{a}}/ \partial \bm{p}^{S_{b}}(t_{j}(\bm{x});\tau)$ in Eq.~(\ref{mini_sup_online2}). When $K\geqslant 2$, we perform mutual mimicry every two of $K$ student networks, leading to the total loss as:
\begin{equation}
\mathcal{L}_{KL\_p}^{S_{1}\sim  S_{K}}=\sum_{1\leqslant a,b \leqslant K}(\mathcal{L}_{KL\_p}^{S_{a}\rightarrow  S_{b}}+\mathcal{L}_{KL\_p}^{S_{b}\rightarrow  S_{a}}).
\end{equation}

\subsubsection{Overall loss} We summarize the above three loss to perform online KD over $K$ student networks:
\begin{equation}
\mathcal{L}^{S_{1}\sim S_{K}}_{OKD}=\mathbb{E}_{\bm{x}\in \mathcal{X}}(\mathcal{L}^{S_{1}\sim S_{K}}_{task}+\mathcal{L}^{S_{1}\sim S_{K}}_{aux\_SSAD}+\mathcal{L}_{KL\_q}^{S_{1}\sim  S_{K}}+\mathcal{L}_{KL\_p}^{S_{1}\sim  S_{K}}).
\label{overall_okd}
\end{equation}
As suggested by DML~\cite{zhang2018deep}, we set the hyper-parameter $\tau=1$ in cross-entropy based task loss and $\tau=3$ in KL-based mimicry loss, and we do not introduce other coefficients about loss weights. Empirically, we found this practice sufficient to work good enough.

\subsubsection{Testing the Student Network}
During the test stage, we can retain any student backbone network from $\{f^{S_{k}}(\cdot)\}_{k=1}^{K}$ and discard all auxiliary branches $\{\{c_{l}^{S_{k}}(\cdot)\}_{l=1}^{L}\}_{k=1}^{K}$, leading to no extra inference costs compared with the original baseline network.

\begin{table*}[tbp]
	\centering
	\caption{Accuracy (\%) of our method for training a single network with auxiliary branches on CIFAR-100. We use four different tasks referred in Section~\ref{four} to train auxiliary branches. During the test stage, we discard all auxiliary branches and only retain the backbone network, leading to no extra inference costs compared with the original baseline network. The numbers in \textbf{Bold} denote the best results. 'Gain(\%)' denotes the performance improvement of our method over baseline.}
	\label{single_net}  
	\begin{tabular}{cccccccc}  
		\toprule
		\multirow{3}{*}{Network}&\multirow{3}{*}{Params}&\multirow{3}{*}{Baseline Eq.~(\ref{ce_cpd})}&\multicolumn{4}{c}{Train a single network with auxiliary branches} & \multirow{3}{*}{Gain(\%)$\uparrow$}\\ 
		\cline{4-7}
		&&&+Supervised&+Self-supervised &+Multi-task &+Self-supervision augmented& \\
		&&&Eq.~(\ref{aux_SCPD})&Eq.~(\ref{aux_SSCPD})& Eq.~(\ref{aux_multi_task})&Eq.~(\ref{ce_ssad})& \\
		\midrule
		WRN-40-2~\cite{zagoruyko2016wide}&2.26M&76.44$_{(\pm 0.20)}$&77.00$_{(\pm 0.23)}$&78.35$_{(\pm 0.20)}$&78.53$_{(\pm 0.23)}$&\textbf{80.90}$_{(\pm 0.22)}$&4.46\\
		
		WRN-40-1~\cite{zagoruyko2016wide}&0.57M&71.95$_{(\pm 0.59)}$&72.36$_{(\pm 0.12)}$&73.58$_{(\pm 0.25)}$&73.38$_{(\pm 0.28)}$&\textbf{75.73}$_{(\pm 0.10)}$&3.78\\
		ResNet-56~\cite{he2016deep}&0.86M&73.00$_{(\pm 0.17)}$ &73.61$_{(\pm 0.14)}$&74.82$_{(\pm 0.23)}$&74.60$_{(\pm 0.26)}$&\textbf{77.52}$_{(\pm 0.43)}$&4.52\\
		ResNet-32$\times$4~\cite{he2016deep}&7.43M&79.56$_{(\pm 0.23)}$&80.09$_{(\pm 0.21)}$&80.53$_{(\pm 0.37)}$&80.57$_{(\pm 0.18)}$&\textbf{83.58}$_{(\pm 0.16)}$&4.02\\
		VGG-13~\cite{simonyan2014very}&9.46M&75.35$_{(\pm 0.21)}$&75.56$_{(\pm 0.20)}$&76.02$_{(\pm 0.34)}$&76.36$_{(\pm 0.23)}$&\textbf{78.56}$_{(\pm 0.09)}$&3.21\\
		
		MobileNetV2~\cite{sandler2018mobilenetv2}&0.81M&73.51$_{(\pm 0.26)}$&74.12$_{(\pm 0.31)}$&74.87$_{(\pm 0.18)}$&75.16$_{(\pm 0.19)}$&\textbf{77.39}$_{(\pm 0.18)}$&3.88\\
		
		ShuffleNetV1~\cite{zhang2018shufflenet}&0.95M&71.74$_{(\pm 0.35)}$ &72.41$_{(\pm 0.28)}$&73.21$_{(\pm 0.32)}$&73.75$_{(\pm 0.25)}$&\textbf{76.19}$_{(\pm 0.14)}$&4.45\\
		
		ShuffleNetV2~\cite{ma2018shufflenet}&1.36M&72.96$_{(\pm 0.33)}$&73.59$_{(\pm 0.17)}$&74.50$_{(\pm 0.33)}$&74.93$_{(\pm 0.12)}$&\textbf{77.82}$_{(\pm 0.13)}$&4.86\\
		\bottomrule
	\end{tabular}
\end{table*}

\section{Experiments}
\subsection{Datasets and Experimental setup}
\paragraph{Image Classification on CIFAR-100} CIFAR-100~\cite{krizhevsky2009learning} is composed of 50K training images and 10K test images in 100 classes. We use the standard data augmentation following~\cite{he2016deep}. We train all networks by the SGD optimizer with a momentum of 0.9, a batch size of 64 and a weight decay of $5\times 10^{-4}$. The initial learning rate starts at 0.05 and is decayed by a factor of 10 at 150, 180 and 210 epochs within the total 240 epochs. For light-weight networks of MobileNetV2, ShuffleNetV1 and ShuffleNetV2, we use an initial learning rate of 0.01.
\paragraph{Image Classification on ImageNet} ImageNet~\cite{deng2009imagenet} is a large-scale image classification dataset composed of 1.2 million training images and 50K validation images in 1000 classes. We use the standard data augmentation following~\cite{he2016deep}. We train all networks by the SGD optimizer with a momentum of 0.9, a batch size of 256 and a weight decay of $1\times 10^{-4}$. The initial learning rate starts at 0.1 and is decayed by a factor of 10 at the 30-th, 60-th and 90-th epochs within the total 100 epochs. 
\paragraph{Transfer Experiments of Image Classification on STL-10 and TinyImageNet} STL-10~\cite{DBLP:journals/jmlr/CoatesNL11} is composed of 5K labelled training images and 8K test images in 10 classes. TinyImageNet\footnote{http://tiny-imagenet.herokuapp.com/} is composed of 100K training images and 10k test images in 200 classes. For data augmentation, we randomly crop the image into the size ranging from $(0.08,1.0)$ of the original size and a random aspect ratio ranging from  $(3/4,4/3)$ of the original aspect ratio. The cropped patch is resized to $32\times 32$ and is randomly horizontal flipped with a probability of 0.5. We train the linear classifiers by the SGD optimizer with a momentum of 0.9, a batch size of 64 and a weight decay of $0$. The initial learning rate starts at 0.1 and is decayed by a factor of 10 at the 30-th, 60-th and 90-th epochs within the total 100 epochs.

\paragraph{Transfer Experiments of Object Detection on Pascal VOC} Following the consistent protocol, we use Pascal VOC~\cite{everingham2010pascal} \texttt{trainval07+12} for training and \texttt{test07} for evaluation. The result set consists of 16551 training images and 4952 test images in 20 classes. The image scale is $800\times 1000$ pixels during training and inference.  We train the model by SGD optimizer with a momentum of 0.9 and a weight decay of $1\times 10^{-4}$. The initial learning rate starts at 0.001 and is decayed by a factor of 10 at the 3-th epoch within the total 4 epochs.

\begin{table*}[t]
	\centering
	\caption{Top-1 accuracy (\%) comparison of SOTA distillation methods across various teacher-student pairs on CIFAR-100. \emph{All results with the form $mean_{(\pm std)}$ are implemented by ourselves with three runs.} The numbers in \textbf{Bold} and \underline{underline} denote the best and the second-best results, respectively. 'Teacher' denotes that we first train the backbone $f^{T}(\cdot)$ and then train auxiliary classifiers $\{c_{l}^{T}(\cdot)\}_{l=1}^{L}$ based on the frozen $f^{T}(\cdot)$. For a fair comparison, all compared methods and 'Ours' are supervised by 'Teacher'. 'Teacher*' denotes that we train $f^{T}(\cdot)$ and $\{c_{l}^{T}(\cdot)\}_{l=1}^{L}$ jointly, leading to a more powerful teacher network. 'Ours*' denotes the results supervised by 'Teacher*' for pursuing better performance.}
	\resizebox{1.\linewidth}{!}{
		\begin{tabular}{cccccccccc}
			\toprule
			Teacher  & WRN-40-2 &WRN-40-2 &ResNet56 &ResNet32$\times$4 & VGG13 &ResNet50  &WRN-40-2&ResNet32$\times$4   \\
			
			Student  & WRN-16-2 &WRN-40-1 &ResNet20 &ResNet8$\times$4 & MobileNetV2 &MobileNetV2  &ShuffleNetV1&ShuffleNetV2\\
			\midrule
			\multirow{2}{*}{Params}& 2.26M & 2.26M &0.86M & 7.43M & 9.46M & 23.71M  &2.26M& 7.43M   \\
			& 0.70M & 0.57M&0.28M & 1.23M & 0.81M &0.81M  &0.95M&1.36M\\
			\midrule
			Teacher  & 76.45 &76.45 &73.44 &79.63 &74.64&79.34&76.45& 79.63\\
			Teacher* & 80.70 &80.70 &77.20 &83.73 &78.48&83.85&80.70& 83.73\\
			
			Student  & 73.57$_{(\pm 0.23)}$ &71.95$_{(\pm 0.59)}$&69.62$_{(\pm 0.26)}$ &72.95$_{(\pm 0.24)}$ &73.51$_{(\pm 0.26)}$&73.51$_{(\pm 0.26)}$&71.74$_{(\pm 0.35)}$&72.96$_{(\pm 0.33)}$\\
			\midrule
			KD~\cite{hinton2015distilling}       & 75.23$_{(\pm 0.23)}$ &73.90$_{(\pm 0.44)}$ &70.91$_{(\pm 0.10)}$ & 73.54$_{(\pm 0.26)}$& 75.21$_{(\pm 0.24)}$&75.80$_{(\pm 0.46)}$& 75.83$_{(\pm 0.18)}$&75.43$_{(\pm 0.33)}$    \\
			FitNet~\cite{romero2014fitnets}& 75.30$_{(\pm 0.42)}$ & 74.30$_{(\pm 0.42)}$  &71.21$_{(\pm 0.16)}$ & 75.37$_{(\pm 0.12)}$&75.42$_{(\pm 0.34)}$&75.41$_{(\pm 0.07)}$& 76.27$_{(\pm 0.18)}$ &76.91$_{(\pm 0.06)}$  \\
			AT~\cite{zagoruyko2016paying}    & 75.64$_{(\pm 0.31)}$  & 74.32$_{(\pm 0.23)}$  & 71.35$_{(\pm 0.09)}$& 75.06$_{(\pm 0.19)}$&74.08$_{(\pm 0.21)}$&76.57$_{(\pm 0.20)}$& 76.51$_{(\pm 0.44)}$&76.32$_{(\pm 0.12)}$   \\
			AB~\cite{heo2019knowledge}& 71.26$_{(\pm 1.32)}$& 74.55$_{(\pm 0.46)}$  &71.56$_{(\pm 0.19)}$ & 74.31$_{(\pm 0.09)}$&74.98$_{(\pm 0.44)}$&75.87$_{(\pm 0.39)}$& 76.43$_{(\pm 0.09)}$&76.40$_{(\pm 0.29)}$\\
			VID~\cite{Sungsoo19Variational}	& 75.31$_{(\pm 0.22)}$ & 74.23$_{(\pm 0.28)}$ &71.35$_{(\pm 0.09)}$ &75.07$_{(\pm 0.35)}$ &75.67$_{(\pm 0.13)}$&75.97$_{(\pm 0.08)}$&76.24$_{(\pm 0.44)}$ &  75.98$_{(\pm 0.41)}$  \\
			RKD~\cite{park2019relational} & 75.33$_{(\pm 0.14)}$ & 73.90$_{(\pm 0.26)}$  &71.67$_{(\pm 0.08)}$ &74.17$_{(\pm 0.22)}$ &75.54$_{(\pm 0.36)}$&76.20$_{(\pm 0.06)}$&75.74$_{(\pm 0.32)}$ & 75.42$_{(\pm 0.25)}$  \\
			SP~\cite{Tung2019Similarity}& 74.35$_{(\pm 0.59)}$ & 72.91$_{(\pm 0.24)}$ &71.45$_{(\pm 0.38)}$ & 75.44$_{(\pm 0.11)}$&75.68$_{(\pm 0.35)}$&76.35$_{(\pm 0.14)}$&76.40$_{(\pm 0.37)}$&76.43$_{(\pm 0.21)}$\\
			CC~\cite{peng2019correlation}& 75.30$_{(\pm 0.03)}$ & 74.46$_{(\pm 0.05)}$ &71.44$_{(\pm 0.10)}$ & 74.40$_{(\pm 0.24)}$&75.66$_{(\pm 0.33)}$&76.05$_{(\pm 0.25)}$&     75.63$_{(\pm 0.30)}$&75.74$_{(\pm 0.18)}$\\
			CRD~\cite{tian2019distillation}& 75.81$_{(\pm 0.33)}$ & 74.76$_{(\pm 0.25)}$ &71.83$_{(\pm 0.42)}$ & 75.77$_{(\pm 0.24)}$&76.13$_{(\pm 0.16)}$&76.89$_{(\pm 0.27)}$&     76.37$_{(\pm 0.23)}$&76.51$_{(\pm 0.09)}$\\
			SSKD~\cite{DBLP:conf/eccv/XuLLL20}& \underline{76.16}$_{(\pm 0.17)}$ & \underline{75.84}$_{(\pm 0.04)}$ & 70.80$_{(\pm 0.02)}$&75.83$_{(\pm 0.29)}$ &76.21$_{(\pm 0.16)}$&\underline{78.21}$_{(\pm 0.16)}$&76.71$_{(\pm 0.31)}$&77.64$_{(\pm 0.24)}$\\
			TOFD~\cite{zhang2020task}& 75.48$_{(\pm 0.21)}$ & 74.45$_{(\pm 0.29)}$ & \underline{72.02}$_{(\pm 0.34)}$&75.45$_{(\pm 0.02)}$ &76.05$_{(\pm 0.16)}$&76.40$_{(\pm 0.13)}$&75.80$_{(\pm 0.20)}$&76.47$_{(\pm 0.22)}$ \\
			SemCKD~\cite{chen2021cross}& 75.02$_{(\pm 0.16)}$ & 73.54$_{(\pm 0.83)}$ & 71.54$_{(\pm 0.11)}$&\underline{75.89}$_{(\pm 0.05)}$ &\underline{76.45}$_{(\pm 0.28)}$&77.23$_{(\pm 0.44)}$&\underline{77.39}$_{(\pm 0.35)}$&\underline{78.26}$_{(\pm 0.24)}$ \\
			\midrule
			HSSAKD (Ours)& 77.20$_{(\pm 0.17)}$  & 77.00$_{(\pm 0.21)}$ &72.58$_{(\pm 0.33)}$ & 77.26$_{(\pm 0.14)}$&77.45$_{(\pm 0.21)}$&78.79$_{(\pm 0.11)}$&  78.51$_{(\pm 0.20)}$&79.93$_{(\pm 0.11)}$\\
			HSSAKD (Ours*)& \textbf{78.67}$_{(\pm 0.20)}$  & \textbf{78.12}$_{(\pm 0.25)}$ &\textbf{73.73}$_{(\pm 0.10)}$ & \textbf{77.69}$_{(\pm 0.05)}$&\textbf{79.27}$_{(\pm 0.12)}$&\textbf{79.43}$_{(\pm 0.24)}$&    \textbf{80.11}$_{(\pm 0.32)}$&\textbf{80.86}$_{(\pm 0.15)}$\\
			\bottomrule
	\end{tabular}}
	\label{cifar}
\end{table*}

\begin{table*}
	\centering
	
	\caption{Top-1 accuracy (\%) comparison of SOTA offline KD methods over ResNet-34-ResNet-18 on ImageNet. }
	\resizebox{1.\linewidth}{!}{
		\begin{tabular}{ccccccccccccc}
			\toprule
			Acc&Teacher&Teacher*  &Student &KD~\cite{hinton2015distilling} &AT~\cite{zagoruyko2016paying} & CC~\cite{peng2019correlation}& SP~\cite{Tung2019Similarity}& RKD~\cite{park2019relational}&CRD~\cite{tian2019distillation} &SSKD~\cite{DBLP:conf/eccv/XuLLL20}&Ours&Ours*\\
			\midrule
		Top-1 &73.31& 75.48 &69.75&70.66 &70.70&69.96&70.62&71.34&71.38&\underline{71.62}&72.16&\textbf{72.39}\\
			Top-5 & 91.42 &92.67 &89.07 &89.88&90.00&89.17&89.80&90.37&90.49&\underline{90.67}&90.85&\textbf{91.00} \\
			
			\bottomrule
	\end{tabular}}
	
	\label{imagenet}
\end{table*}

\paragraph{Auxiliary self-supervised classification task} Unless otherwise specified, we employ 4-way rotations from $\{0^{\circ},90^{\circ},180^{\circ},270^{\circ}\}$ as an auxiliary self-supervised task, as illustrated in Fig.~\ref{Overview_methods}.

\subsection{Experiments on CIFAR-100 and ImageNet Classification}

\begin{table*}[tbp]
	\centering
	\caption{Top-1 accuracy(\%) comparisons of our method against SOTA online mutual knowledge distillation methods on CIFAR-100 and ImageNet. \emph{All results with the form $mean_{(\pm std)}$ are implemented by ourselves with three runs.} The numbers in \textbf{Bold} and \underline{underline} denote the best and the second-best results, respectively. 'Gain(\%)' denotes the performance improvement of our method over the second-best result.}
	\label{online_kd}  
	\begin{tabular}{ccccccccc}  
		\toprule
		Network&Baseline&DML~\cite{zhang2018deep}&ONE~\cite{zhu2018knowledge}&OKDDip~\cite{chen2020online}&KDCL~\cite{guo2020online}&AFD~\cite{chung2020feature}&HSSAKD (Ours) & Gain(\%)$\uparrow$\\  
		\midrule
		WRN-40-2&76.44$_{(\pm 0.20)}$&78.96$_{(\pm 0.13)}$&78.76$_{(\pm 0.14)}$&79.23$_{(\pm 0.34)}$&78.50$_{(\pm 0.29)}$&\underline{79.54}$_{(\pm 0.13)}$&\textbf{82.58$_{(\pm 0.21)}$}&3.04\\
		
		WRN-40-1&71.95$_{(\pm 0.59)}$&74.33$_{(\pm 0.12)}$&74.56$_{(\pm 0.25)}$&74.89$_{(\pm 0.29)}$&74.20$_{(\pm 0.22)}$&\underline{75.23}$_{(\pm 0.05)}$&\textbf{76.67$_{(\pm 0.41)}$}&1.44\\
		ResNet-56&73.00$_{(\pm 0.17)}$&75.40$_{(\pm 0.24)}$&75.64$_{(\pm 0.12)}$&76.25$_{(\pm 0.38)}$&75.28$_{(\pm 0.13)}$&\underline{76.78}$_{(\pm 0.28)}$&\textbf{78.16$_{(\pm 0.56)}$}&1.38\\
		
		ResNet-32$\times$4&79.56$_{(\pm 0.23)}$&81.61$_{(\pm 0.27)}$&82.34$_{(\pm 0.19)}$&81.21$_{(\pm 0.10)}$&80.76$_{(\pm 0.07)}$&\underline{82.91}$_{(\pm 0.17)}$&\textbf{84.91$_{(\pm 0.19)}$}&2.00\\
		
		VGG-13&75.35$_{(\pm 0.21)}$&77.77$_{(\pm 0.19)}$&78.45$_{(\pm 0.34)}$&77.32$_{(\pm 0.09)}$&77.22$_{(\pm 0.16)}$&\underline{78.61}$_{(\pm 0.09)}$&\textbf{80.44$_{(\pm 0.05)}$}&1.83\\
		
		MobileNetV2&73.51$_{(\pm 0.26)}$&76.57$_{(\pm 0.16)}$&76.13$_{(\pm 0.17)}$&76.79$_{(\pm 0.41)}$&75.63$_{(\pm 0.20)}$&\underline{77.05}$_{(\pm 0.37)}$&\textbf{78.85$_{(\pm 0.13)}$}&1.80\\
		
		ShuffleNetV1&71.74$_{(\pm 0.35)}$&75.01$_{(\pm 0.42)}$&74.63$_{(\pm 0.08)}$&75.41$_{(\pm 0.29)}$&74.82$_{(\pm 0.59)}$&\underline{75.82}$_{(\pm 0.34)}$&\textbf{78.34$_{(\pm 0.03)}$}&2.52\\
		
		ShuffleNetV2&72.96$_{(\pm 0.33)}$&76.41$_{(\pm 0.18)}$&75.74$_{(\pm 0.20)}$&76.29$_{(\pm 0.22)}$&75.10$_{(\pm 0.35)}$&\underline{77.21}$_{(\pm 0.14)}$&\textbf{79.98$_{(\pm 0.12)}$}&2.77\\
		\bottomrule
	\end{tabular}
\end{table*}

\begin{table*}
	\centering
	
	\caption{Top-1 accuracy (\%) comparison of SOTA online KD methods on ImageNet.}
	\begin{tabular}{cccccccc}
		\toprule
		Network & Baseline&DML~\cite{zhang2018deep}&ONE~\cite{zhu2018knowledge}&OKDDip~\cite{chen2020online}&KDCL~\cite{guo2020online}&AFD~\cite{chung2020feature}&HSSAKD (Ours)\\
		\midrule
		ResNet-18 & 69.75 &70.48 &70.55 &70.73 &\underline{70.83} &70.51 &\textbf{71.49} \\
		
		\bottomrule
	\end{tabular}
	
	\label{online_imagenet}
\end{table*}

\begin{table*}
	\caption{Linear classification accuracy (\%) of transfer learning on the student MobileNetV2 pre-trained using the teacher VGG-13. The numbers in \textbf{Bold} and \underline{underline} denote the best and the second-best results, respectively.}
	\centering
	\begin{tabular}{lccccccccccccc}
		\toprule
		Transferred Dataset & Teacher & Student &KD &FitNet&AT &AB& VID& RKD&SP& CC&CRD &SSKD&Ours  \\
		\midrule
		CIFAR-100$\rightarrow $ STL-10 & 63.08&67.76  &67.90 &69.41&67.37&67.82&69.29&69.74&68.96&69.13&70.09&\underline{71.03}&\textbf{74.66} \\
		CIFAR-100$\rightarrow $ TinyImageNet & 29.53& 34.69 &34.15&36.04&34.44&34.79&36.09&37.21&35.69&36.43&38.17&\underline{39.07}&\textbf{42.57} \\
		\bottomrule
	\end{tabular}
	
	\label{transfer}
\end{table*}

\begin{table*}
	\centering
	\caption{Comparison of detection AP (\%) (Average Precision) on individual classes and mAP (mean Average Precision) (\%) on Pascal VOC~\cite{everingham2010pascal} using ResNet-18 as the backbone pre-trained by various KD methods over the Faster-RCNN system~\cite{ren2016faster}. The numbers in \textbf{Bold} and \underline{underline} denote the best and the second-best results, respectively.}
	\resizebox{1.\linewidth}{!}{
		\begin{tabular}{lccccccccccccccccccccc}
			\toprule
			\multirow{2}{*}{Method} & \multirow{2}{*}{mAP} &\multicolumn{20}{c}{AP (Average Precision)}  \\ \cmidrule{3-22}
			&  &areo &bike&brid &boat& bottle& bus&car& cat&chair &cow&table&dog&horse&mbike&person&plant&sheep&sofa&train&tv  \\
			\midrule
			Baseline &76.2&\underline{79.0}&83.4&77.5&63.4&65.0&80.3&85.5&86.7&59.4&81.7&69.1&84.2&84.2&81.3&83.7&48.3&80.0&73.6&82.6&74.6 \\
			
			KD~\cite{hinton2015distilling} &77.1&78.6&\underline{84.4}&78.0&\underline{68.0}&62.8&\underline{82.9}&85.9&\textbf{88.4}&60.7&81.5&68.6&\underline{85.9}&84.4&82.7&\underline{84.5}&48.8&79.5&74.8&\textbf{85.0}&75.9 \\
			
			CRD~\cite{tian2019distillation} &77.4&77.5&\textbf{84.9}&77.3&66.3&\underline{65.6}&82.3&\underline{86.6}&88.2&62.1&81.8&\underline{70.8}&85.7&\underline{84.9}&\underline{82.8}&84.3&51.6&\underline{81.1}&\underline{75.2}&83.1&75.2 \\
			
			SSKD~\cite{DBLP:conf/eccv/XuLLL20}& \underline{77.6}&78.6&84.2&\underline{78.2}&66.2&63.3&82.8&86.1&87.3&\textbf{63.4}&\textbf{84.4}&\textbf{71.8}&84.5&\textbf{85.1}&\textbf{83.1}&83.9&\underline{51.9}&80.1&\textbf{77.2}&84.1&\underline{76.0} \\
			
			HSSAKD &\textbf{78.4}& \textbf{84.8} &\textbf{84.9}&\textbf{81.8}&\textbf{68.2}&\textbf{66.9}&\textbf{84.1}&\textbf{86.7}&\underline{88.2}&\underline{62.1}&\underline{82.6}&70.3&\textbf{86.4}&84.3&82.3&\textbf{84.8}&\textbf{53.2}&\textbf{81.3}&74.8&\underline{84.2}&\textbf{78.2} \\
			\bottomrule
	\end{tabular}}
	
	\label{detection}
\end{table*}
\subsubsection{Training a Single Network with Auxiliary Branches}
\label{training_single}
 Initially, we examine the efficacy of four different auxiliary tasks referred in Section~\ref{four} for enhancing the representation learning over a single network for solving the primary supervised task. As illustrated in Fig.~\ref{SMG}, we guide all auxiliary branches to learn additional tasks. The results about four different auxiliary tasks over the baseline are shown in Table~\ref{single_net}.

 Although forcing auxiliary branches to learn various tasks can consistently improve performance over the original baseline network, using our proposed self-supervision augmented task as an auxiliary task can result in maximum accuracy gains. Moreover, we find that using a supervised task as the same as the primary task only leads to marginal improvements with an average gain of 0.53\% upon the baseline across various networks. This is because the homogeneous supervised task may not contribute much extra knowledge. Somewhat surprisingly, using a self-supervised task that is orthogonal to the primary task can lead to moderate improvements with an average gain of 1.60\%. This is possibly because this self-supervised task can bring extra self-supervision knowledge and explicitly facilitate feature representation learning. We further find that incorporating supervised and self-supervised task into a multi-task framework only achieves comparable performance with the self-supervised counterpart. 
 
 In contrast, our self-supervision augmented task leads to significant improvements with an average gain of 4.15\%. The results suggest that the self-supervision augmented task can explore more meaningful joint knowledge to encourage better feature representations than any single task or multi-task learning. 

\subsubsection{Teacher-Student Based Offline HSSAKD}  Based on the success of the self-supervision augmented task, we consider distilling self-supervision augmented distributions as new knowledge from hierarchical feature maps between teacher and student networks. As shown in Table~\ref{cifar} and Table~\ref{imagenet}, we compare our HSSAKD with other popular KD methods across widely used network architectures on CIFAR-100 and ImageNet. Our HSSAKD achieves the best performance among all KD methods. It is noteworthy that HSSAKD can significantly outperform the previous SOTA method SSKD~\cite{DBLP:conf/eccv/XuLLL20} by an average accuracy gain of 2.56\% on CIFAR-100 and a top-1 gain of 0.77\% on ImageNet. The results verify that encoding the self-supervised auxiliary task as an augmented distribution in our HASKD is better than self-supervised contrastive distributions in SSKD to perform KD-based representation learning.

\subsubsection{Multi-Student Based Online HSSAKD} Based on the success in offline HSSAKD, it is also interesting to extend our HSSAKD framework to the scenario of online KD for mutually transferring self-supervision augmented distributions. As shown in Table~\ref{online_kd}, we compare HSSAKD with other representative online KD methods across widely used network architectures on CIFAR-100 and ImageNet. For fair comparisons, all methods use two identical peer networks (\emph{i.e.} $K=2$) for online KD. Our HSSAKD achieves the best performance among all online KD methods. Specifically, HSSAKD can significantly outperforms the seminal DML~\cite{zhang2018deep} and previous SOTA method AFD~\cite{chung2020feature} by an average accuracy gain of 2.98\% and 2.20\% on CIFAR-100, respectively. Further experiments on the large-scale ImageNet for collaboratively training two ResNet-18 verify the superiority of our HSSAKD. We remark that previous online KD methods, such as DML~\cite{zhang2018deep}, ONE~\cite{zhu2018knowledge} and KDCL~\cite{guo2020online}, often focus on distilling the conventional supervised class probability distributions and only differ in the mechanisms for distillation or producing an online teacher. In contrast, our HSSAKD goes on a step further to introduce a new type of knowledge of self-supervision augmented distributions, which achieves promising results in the field of online KD.

\subsection{Transfer Experiments of Learned Feature Representations}
\subsubsection{Transfer Experiments on STL-10 and TinyImageNet for Image Classification} Beyond the accuracy on the object dataset, we also expect that the student network can produce the generalized feature representations that transfer well to other unseen semantic recognition datasets. To this end, we freeze the feature extractor pre-trained on the upstream CIFAR-100, and then train two linear classifiers based on frozen pooled features for downstream STL-10 and TinyImageNet respectively, following the common linear classification protocol~\cite{tian2019distillation}. As shown in Table~\ref{transfer}, we can observe that both SSKD and HSSAKD achieve better accuracy than other comparative methods, demonstrating that using self-supervision auxiliary tasks for distillation is conducive to generating better feature representations. Moreover, HSSAKD can significantly outperform the best-competing SSKD by an accuracy gain of 3.63\% on STL-10 and an accuracy gain of 3.50\% on TinyImageNet.

\paragraph{Transfer Experiments on Pascal VOC for Object Detection} We further evaluate the student network ResNet-18 pre-trained with the teacher ResNet-34 on ImageNet as a backbone to carry out downstream object detection on Pascal VOC~\cite{everingham2010pascal}. We use Faster-RCNN~\cite{ren2016faster} framework and follow the standard data preprocessing and finetuning strategy. The comparison on detection performance towards AP (Average Precision) on individual classes and mAP (mean Average Precision) is shown in Table~\ref{detection}.  Our method outperforms the original baseline by 2.27\% mAP and the best-competing SSKD by 0.85\% mAP. Moreover, our method can also achieve the best or second-best AP in most classes compared with other widely used KD methods. These results verify that our method can guide a network to learn better feature representations for semantic recognition tasks.

\begin{figure}[tbp]  
	\centering  
	\includegraphics[width=0.8\linewidth]{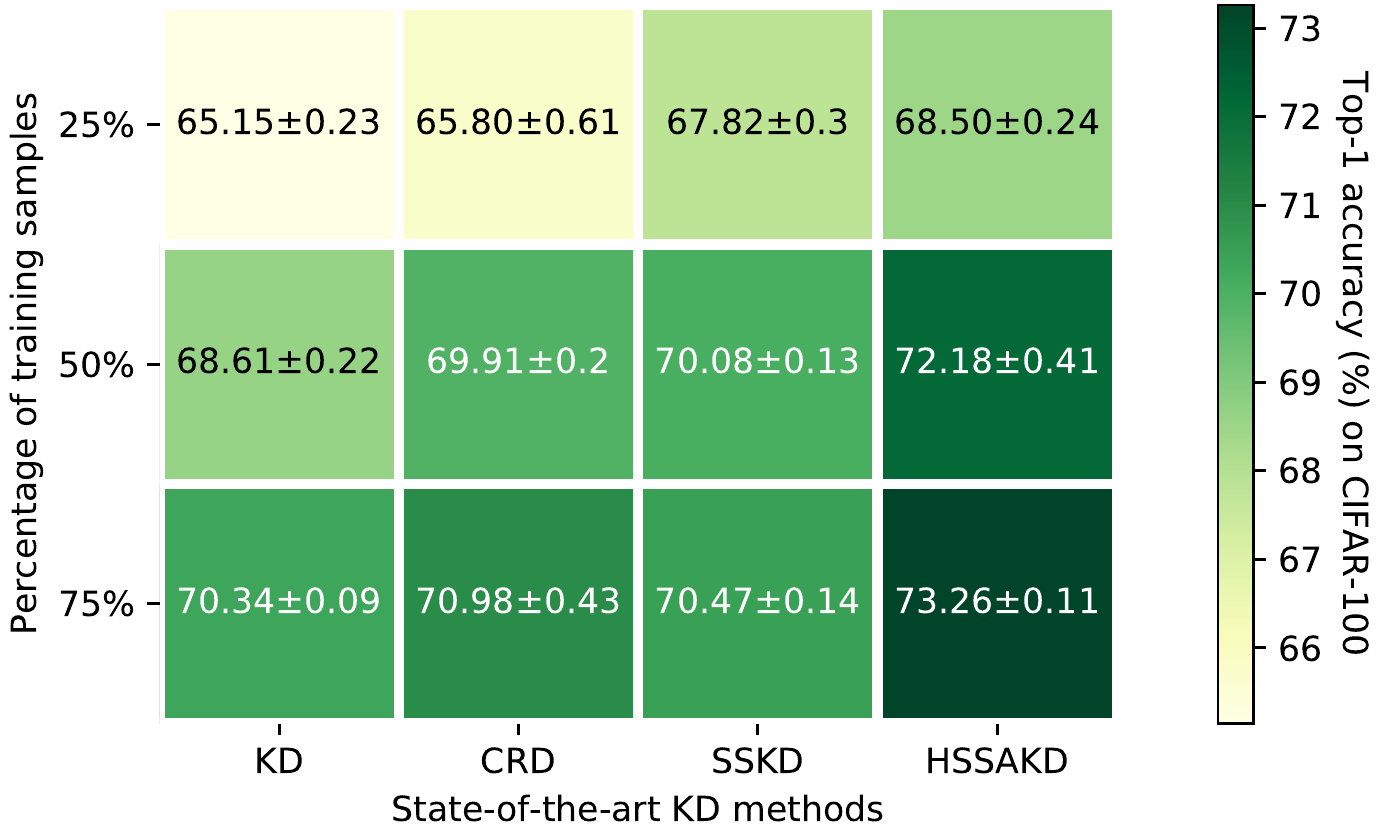}
	\caption{Top-1 accuracy (\%) comparison on CIFAR-100 under few-shot scenario with various percentages of training samples. We use the ResNet56-ResNet20 as the teacher-student pair for evaluation.}  
	\label{few_shot}
\end{figure}

\subsection{Efficacy under Few-shot Scenario} We compare our method with conventional KD and SOTA CRD and SSKD under few-shot scenarios by retaining 25\%, 50\% and 75\% training samples. For a fair comparison, we use the same data split with a stratified sampling strategy for each few-shot setting while maintaining the original test set. As shown in Fig.~\ref{few_shot}, our method can consistently surpass other offline KD methods by large margins under various few-shot settings. Moreover, it is noteworthy that by using only 25\% training samples, our method can achieve comparable accuracy with the baseline trained on the complete set. This is because our method can effectively learn general feature representations from limited data. In contrast, previous methods often focus on mimicking the inductive biases from intermediate feature maps or cross-sample relationships, which may overfit the limited set and generalize worse to the test set.

\section{Ablation Study and Analysis}

\begin{table}
	\caption{Top-1 accuracy (\%) comparison towards ablation study of loss terms on the student networks of WRN-16-2 and ShuffleNetV1 supervised through the pre-trained teacher network WRN-40-2 for \emph{offline KD} on CIFAR-100.}
	\centering
	\begin{tabular}{lcccccccccccc}
		\toprule
		Loss terms & WRN-16-2 &ShuffleNetV1  \\
		\midrule
		$\mathcal{L}_{task}^{S}$ & 73.57$_{(\pm 0.23)}$  &71.74$_{(\pm 0.35)}$ \\
		$\mathcal{L}_{task}^{S}+\mathcal{L}_{KL\_q}^{S}$ &77.11$_{(\pm 0.13)}$ & 78.87$_{(\pm 0.25)}$ \\
		$\mathcal{L}_{task}^{S}+\mathcal{L}_{KL\_q}^{S}+$KD~\cite{hinton2015distilling} & 78.23$_{(\pm 0.23)}$ & 79.06$_{(\pm 0.30)}$\\
		$\mathcal{L}_{task}^{S}+\mathcal{L}_{KL\_q}^{S}+\mathcal{L}_{KL\_p}^{S}$ & \textbf{78.67}$_{(\pm 0.20)}$ & \textbf{80.11}$_{(\pm 0.32)}$ \\
		\bottomrule
	\end{tabular}
	\label{offline_loss_terms}
\end{table}
\begin{table}
	\caption{Top-1 accuracy (\%) comparison towards ablation study of loss terms on the student networks of WRN-16-2 and ShuffleNetV1 for \emph{online mutual KD} on CIFAR-100.}
	\centering
	\resizebox{1.\linewidth}{!}{
	\begin{tabular}{lcccccccccccc}
		\toprule
		Loss terms & WRN-16-2 &ShuffleNetV1  \\
		\midrule
		$\mathcal{L}_{task}^{S_{1}\sim S_{2}}$ & 73.57$_{(\pm 0.23)}$  &71.74$_{(\pm 0.35)}$ \\
		$\mathcal{L}_{task}^{S_{1}\sim S_{2}}+\mathcal{L}_{aux\_SSAD}^{S_{1}\sim S_{2}}$ &76.15$_{(\pm 0.31)}$ & 76.19$_{(\pm 0.14)}$ \\
		$\mathcal{L}_{task}^{S_{1}\sim S_{2}}+\mathcal{L}_{aux\_SSAD}^{S_{1}\sim S_{2}}+\mathcal{L}_{KL\_q}^{S_{1}\sim S_{2}}$ &76.83$_{(\pm 0.29)}$ & 77.57$_{(\pm 0.21)}$ \\
		$\mathcal{L}_{task}^{S_{1}\sim S_{2}}+\mathcal{L}_{aux\_SSAD}^{S_{1}\sim S_{2}}+\mathcal{L}_{KL\_q}^{S_{1}\sim S_{2}}+$DML~\cite{zhang2018deep} & 76.89$_{(\pm 0.35)}$ & 77.73$_{(\pm 0.17)}$\\
		$\mathcal{L}_{task}^{S_{1}\sim S_{2}}+\mathcal{L}_{aux\_SSAD}^{S_{1}\sim S_{2}}+\mathcal{L}_{KL\_q}^{S_{1}\sim S_{2}}+\mathcal{L}_{KL\_p}^{S_{1}\sim S_{2}}$ & \textbf{77.30}$_{(\pm 0.15)}$ & \textbf{78.34}$_{(\pm 0.03)}$ \\
		\bottomrule
	\end{tabular}}
	\label{online_loss_terms}
\end{table}
 \begin{figure*}[tbp]  
	\centering  
	\includegraphics[width=1\linewidth]{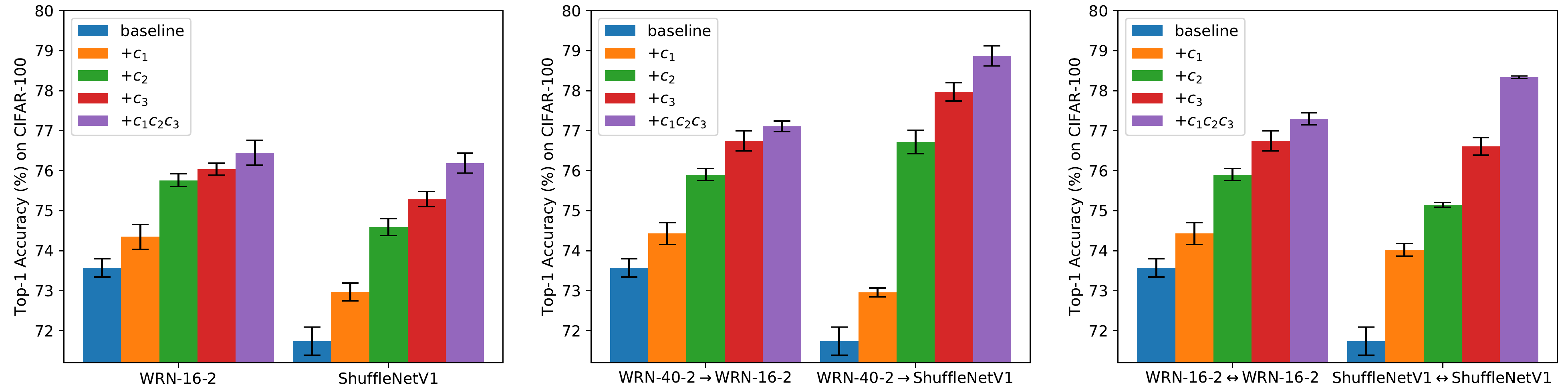}
	\caption{Ablation study of the effectiveness of various auxiliary branches for learning and distilling self-supervision augmented knowledge on CIFAR-100 under: (1) \emph{\textbf{left}}: train a single network with auxiliary branches; (2) \emph{\textbf{middle}}: conventional offline HSSAKD; (3) \emph{\textbf{right}}: online mutual HSSAKD.}  
	\label{ablation_aux_cls}
\end{figure*}
\subsection{Effect of loss terms}
\label{kd_loss}
 To examine each component of our proposed HSSAKD, we compare three variants under the conventional offline and online scenarios settings, respectively. As shown in Table~\ref{offline_loss_terms}, for offline HSSAKD, three variants are (1) our core loss $\mathcal{L}^{S}_{KL\_q}$ for distilling self-supervision augmented distributions; (2) $\mathcal{L}^{S}_{KL\_q}$ +conventional KD~\cite{hinton2015distilling}; (3) $\mathcal{L}^{S}_{KL\_q}+\mathcal{L}^{S}_{KL\_p}$. The difference between the conventional KD and our $\mathcal{L}_{KL\_p}$ is that the latter also distills supervised class posterior from \emph{extra self-supervision transformed images} from teacher to student. As shown in Table~\ref{online_loss_terms}, for online HSSAKD, three variants are (1) our core loss $\mathcal{L}^{S_{1}\sim S_{2}}_{aux\_SSAD}+\mathcal{L}^{S_{1}\sim S_{2}}_{KL\_q}$ for online learning and distilling self-supervision augmented distributions; (2) $\mathcal{L}^{S_{1}\sim S_{2}}_{aux\_SSAD}+\mathcal{L}^{S_{1}\sim S_{2}}_{KL\_q}$+DML~\cite{zhang2018deep}; (3) $\mathcal{L}^{S_{1}\sim S_{2}}_{aux\_SSAD}+\mathcal{L}^{S_{1}\sim S_{2}}_{KL\_q}+\mathcal{L}^{S_{1}\sim S_{2}}_{KL\_p}$. The difference between DML and our $\mathcal{L}^{S_{1}\sim S_{2}}_{KL\_p}$ is that the latter also performs mutual distillation of supervised class posterior from \emph{extra self-supervision transformed images} between two online networks.

 We can derive similar observations under these two scenarios. First, learning and distilling hierarchical self-supervision augmented knowledge through multiple auxiliary branches by loss $\mathcal{L}^{S}_{KL\_q}$ for offline HSSAKD and loss $\mathcal{L}_{aux\_SSAD}^{S_{1}\sim S_{2}}+\mathcal{L}_{KL\_q}^{S_{1}\sim S_{2}}$ for online HSSAKD substantially boosts the accuracy upon the original task loss $\mathcal{L}_{task}$. Moreover, we compare offline $\mathcal{L}^{S}_{KL\_p}$ with conventional KD, as well as online $\mathcal{L}^{S_{1}\sim S_{2}}_{KL\_p}$ with DML. The superior results over previous methods suggest that distilling probabilistic class posterior from those self-supervised transformed images is also beneficial to KD-based feature learning. 
 

\subsection{Effect of auxiliary branches} We propose to append several auxiliary branches to a network over various depths to learn and transfer diverse self-supervision augmented distributions extracted from hierarchical features. To examine the effectiveness of each auxiliary branch, we individually evaluate each one and drop others. As shown in Fig.~\ref{ablation_aux_cls}, we can observe that each auxiliary branch is beneficial to performance improvements. Moreover, an auxiliary branch attached in the deeper layer often achieves a more accuracy gain than one in the shallower layer. This can be attributed to more informative semantic knowledge encoded in highly abstract features output from deep layers. Finally, we can aggregate all auxiliary branches to maximize accuracy gains, suggesting that it is meaningful to learn and transfer self-supervision augmented knowledge from \emph{hierarchical feature maps}.

\begin{table}
	\caption{Top-1 accuracy (\%) comparison towards various distributions as knowledge for \emph{offline KD} on the student networks of WRN-16-2 and ShuffleNetV1 supervised through the pre-trained teacher network WRN-40-2 on CIFAR-100.}
	\label{offline}
	\centering
	\resizebox{1.\linewidth}{!}{
	\begin{tabular}{lcccccccccccc}
		\toprule
		Knowledge for KD & WRN-16-2 &ShuffleNetV1  \\
		\midrule
		None & 73.57$_{(\pm 0.23)}$  &71.74$_{(\pm 0.35)}$ \\
		Supervised $\bm{p}\in \mathbb{R}^{N}$ & 74.73$_{(\pm 0.33)}$  &73.24$_{(\pm 0.27)}$ \\
		Self-supervised $\bm{\mu}\in \mathbb{R}^{M}$ &75.22$_{(\pm 0.13)}$ & 73.62$_{(\pm 0.15)}$ \\
		Multi-task $\bm{p}\in \mathbb{R}^{N}$ and $\bm{\mu}\in \mathbb{R}^{M}$  & 74.94$_{(\pm 0.18)}$ & 74.10$_{(\pm 0.26)}$\\
		Self-supervision augmented $\bm{q}\in \mathbb{R}^{N*M}$ & \textbf{78.67}$_{(\pm 0.20)}$ & \textbf{80.11}$_{(\pm 0.32)}$ \\
		\bottomrule
	\end{tabular}}
\end{table}
\begin{figure*}
	\centering 
	\begin{subfigure}[t]{0.18\textwidth}
		\centering
		\includegraphics[width=\textwidth]{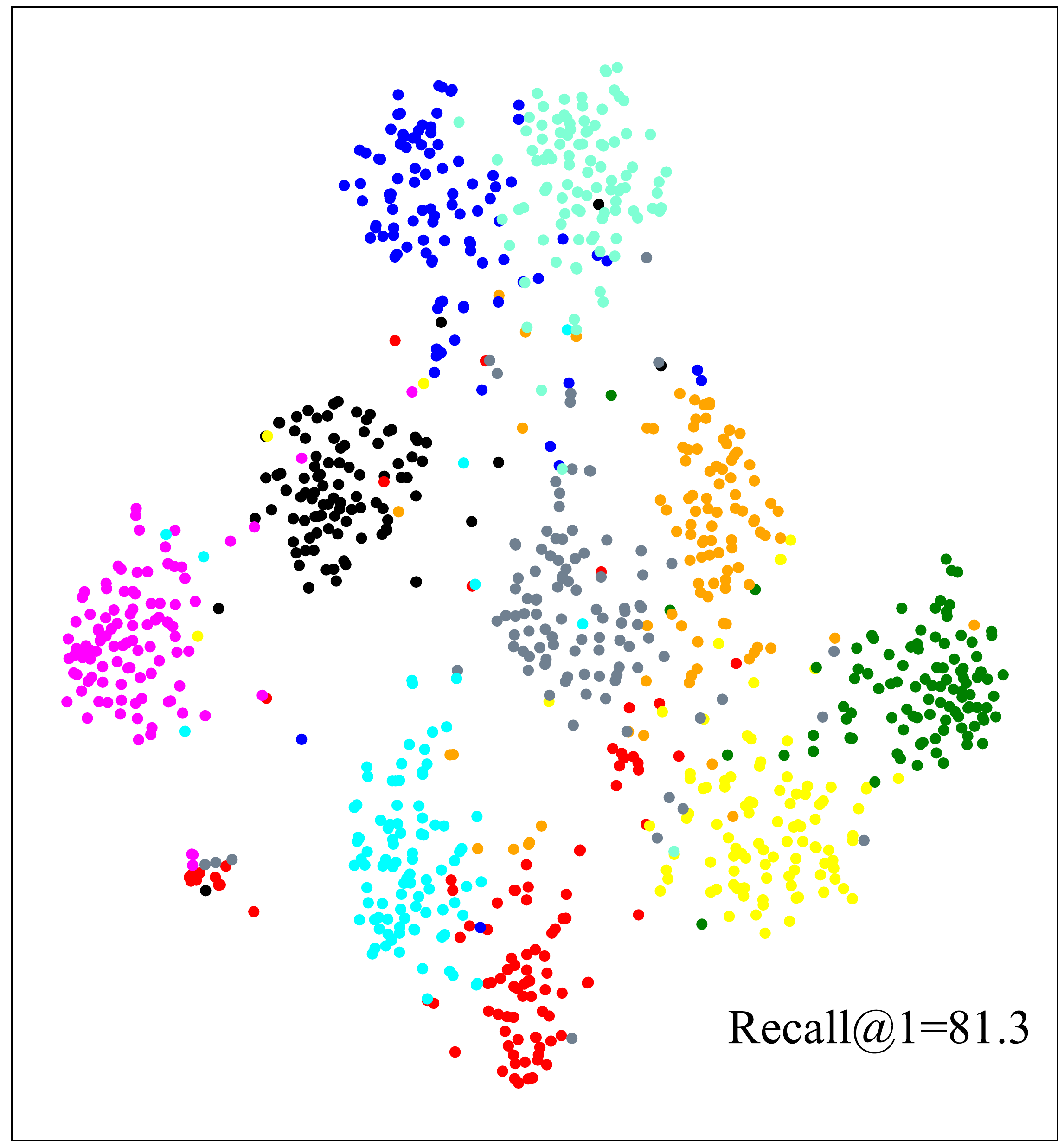}
		\caption{Baseline}
		\label{tsne_baseline}
	\end{subfigure}
	\begin{subfigure}[t]{0.18\textwidth}
		\centering
		\includegraphics[width=\textwidth]{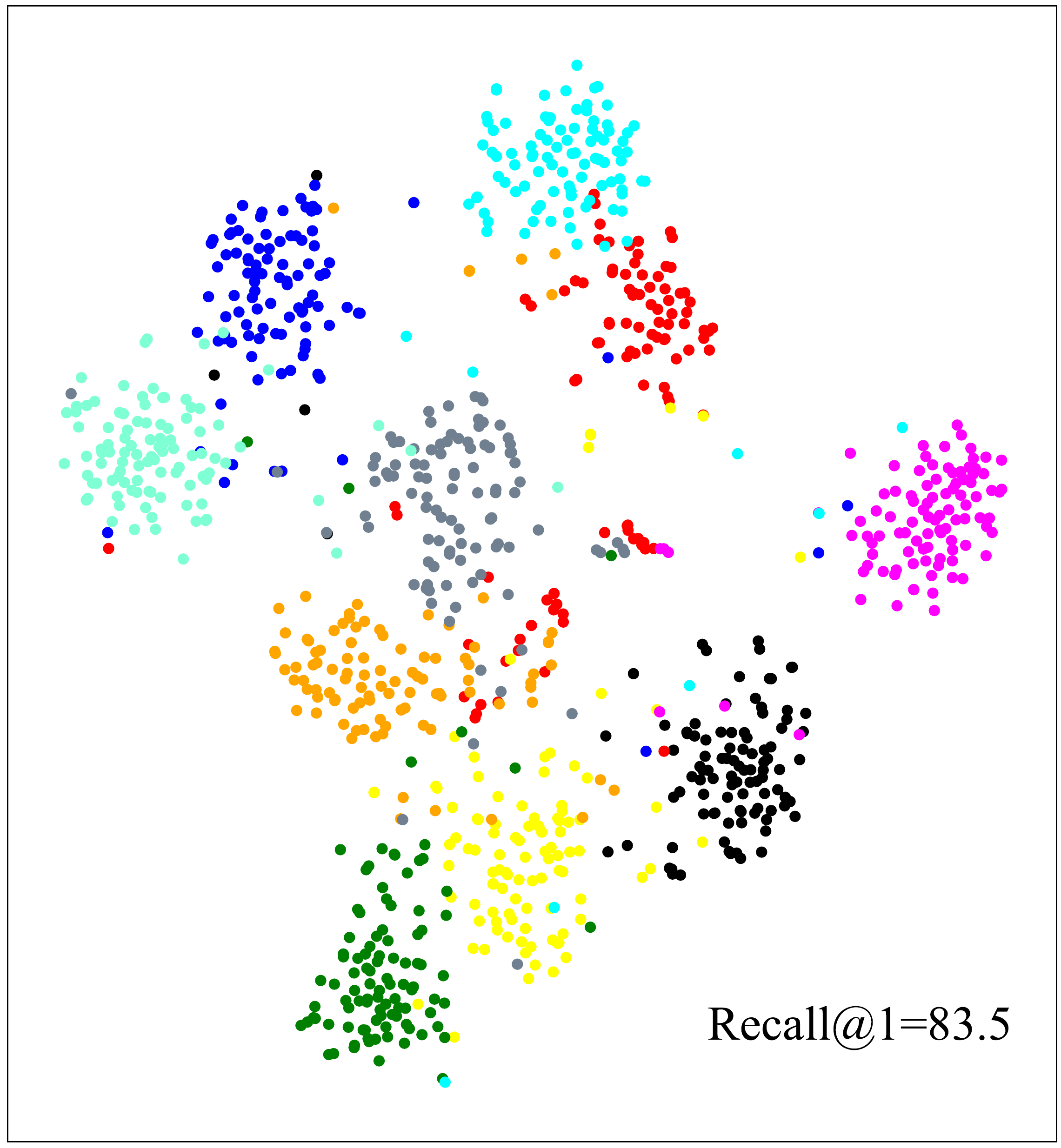}
		\caption{KD~\cite{hinton2015distilling}}
		\label{tsne_kd}
	\end{subfigure}
	\begin{subfigure}[t]{0.18\textwidth}
		\centering
		\includegraphics[width=\textwidth]{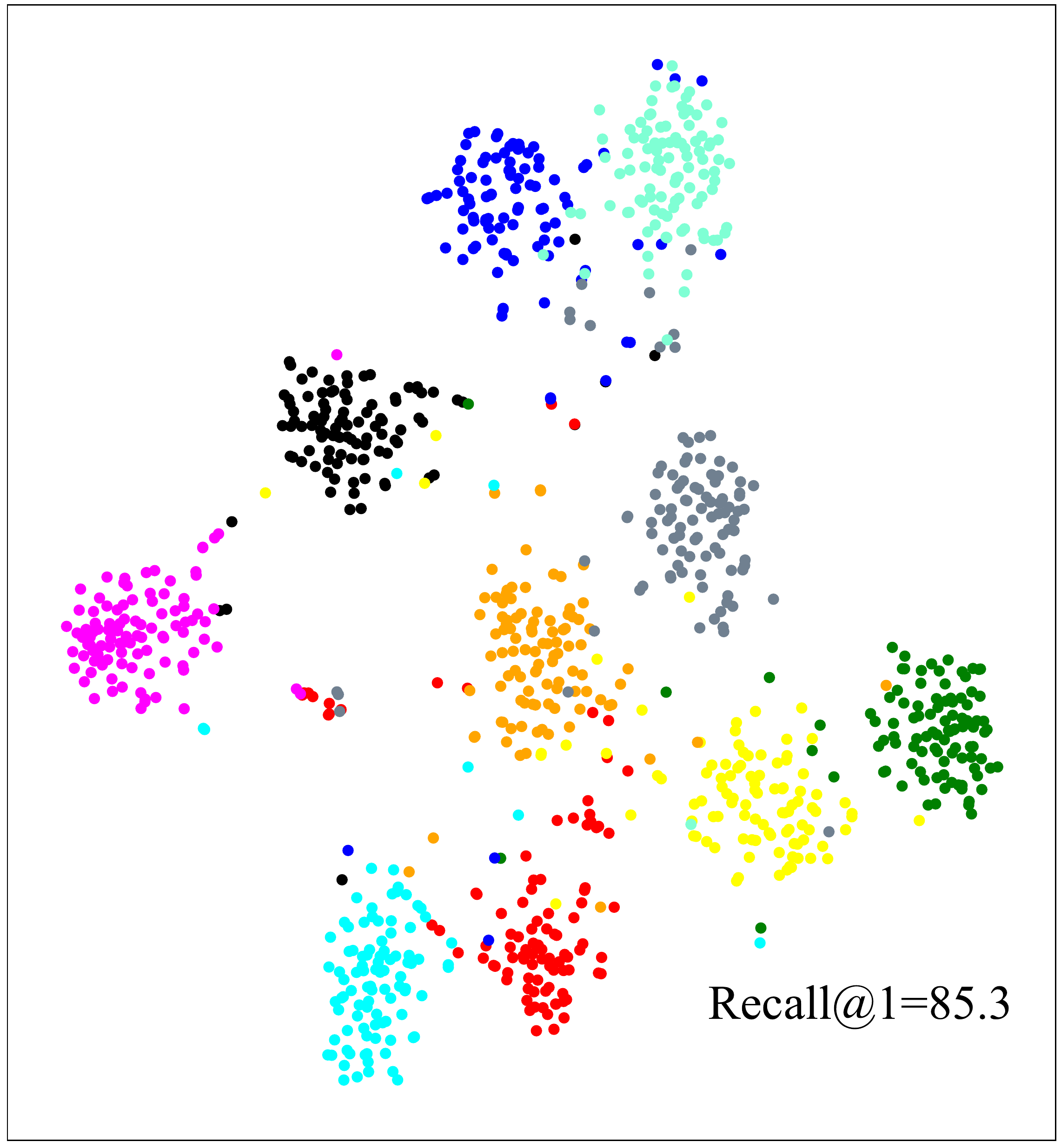}
		\caption{CRD~\cite{tian2019distillation}}
		\label{tsne_crd}
	\end{subfigure}
	\begin{subfigure}[t]{0.18\textwidth}
		\centering
		\includegraphics[width=\textwidth]{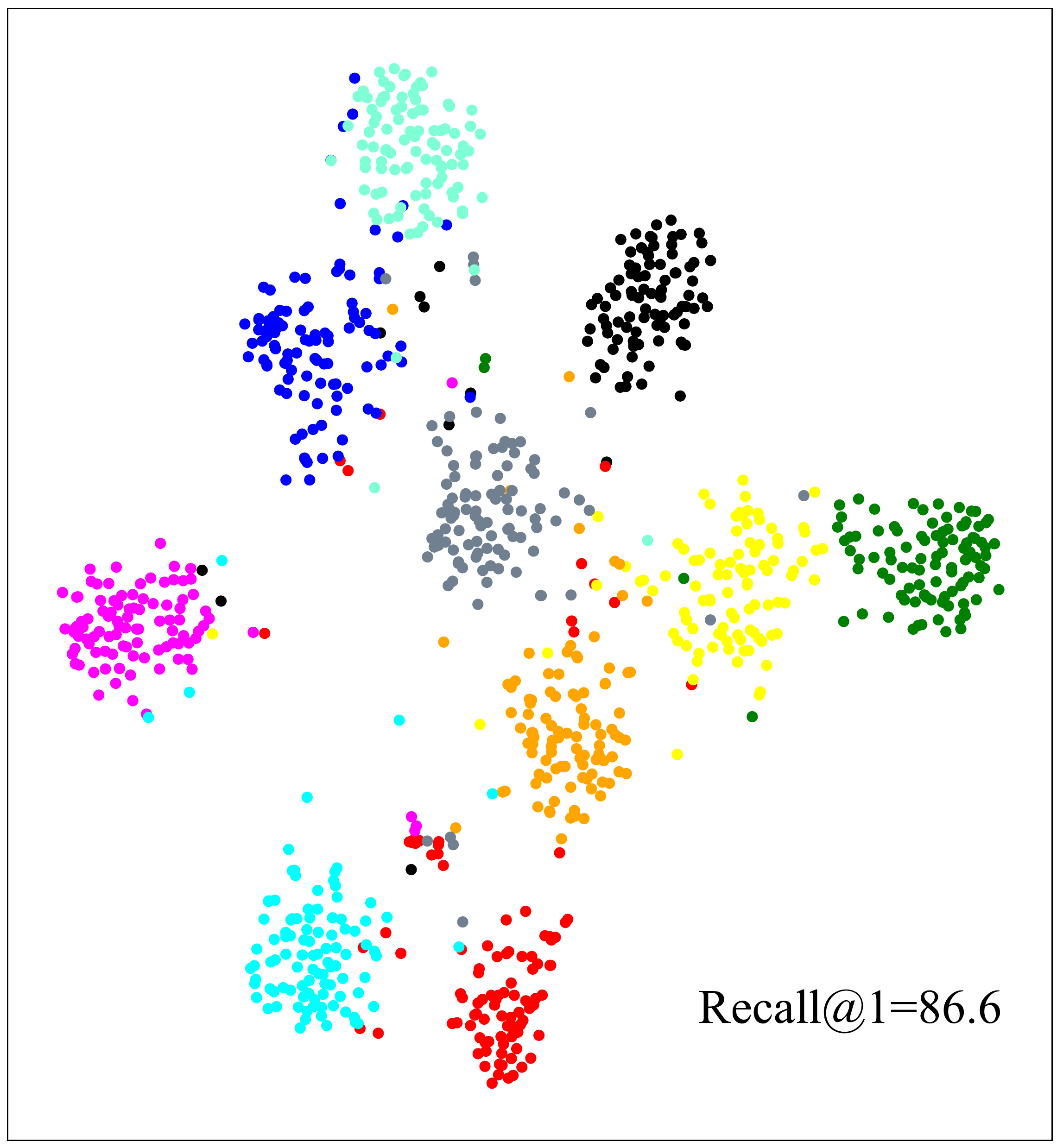}
		\caption{SSKD~\cite{DBLP:conf/eccv/XuLLL20}}
		\label{tsne_sskd}
	\end{subfigure}
	\begin{subfigure}[t]{0.18\textwidth}
		\centering
		\includegraphics[width=\textwidth]{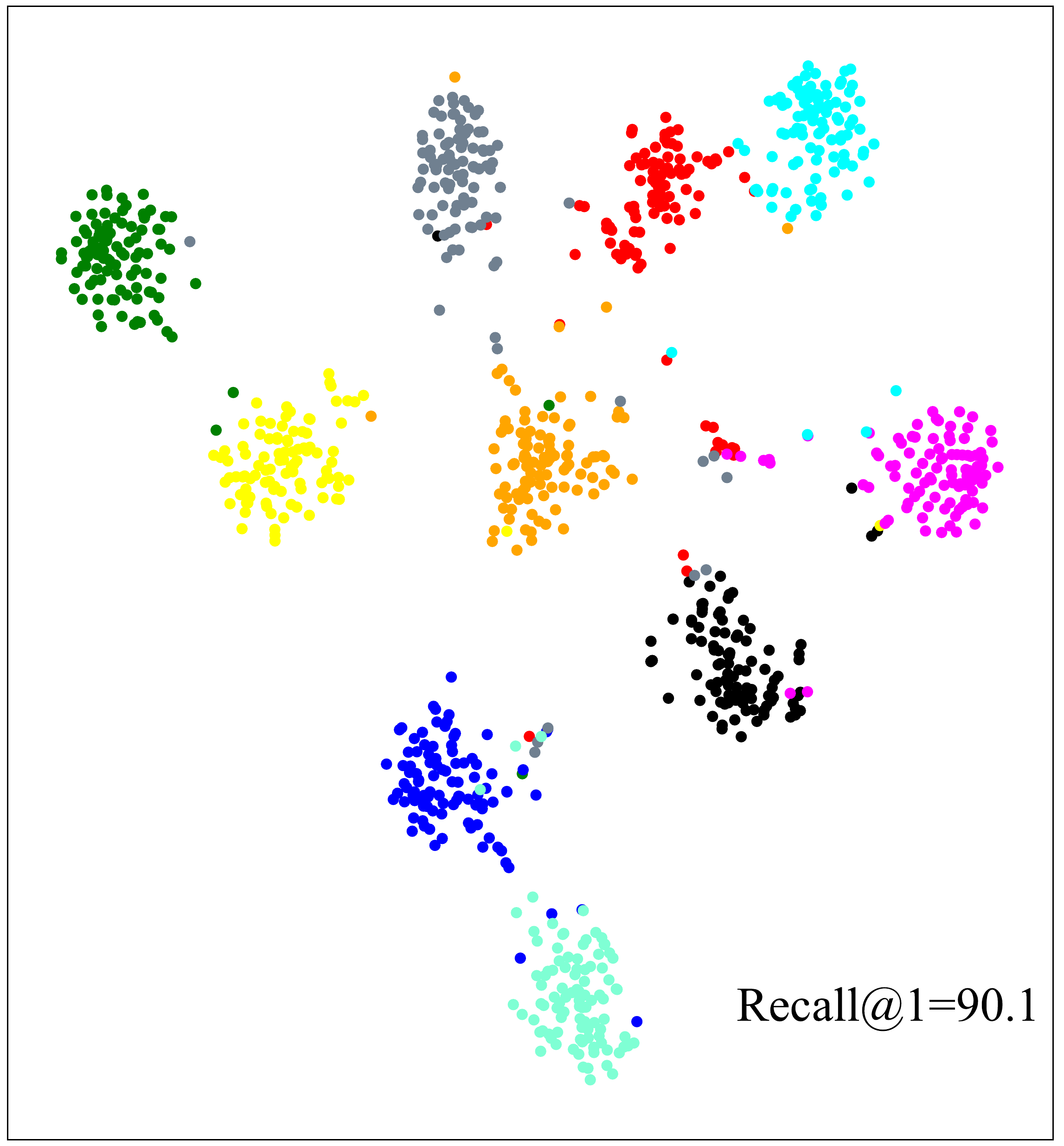}
		\caption{HSSAKD (Ours)}
		\label{tsne_hssakd}
	\end{subfigure}
	
	\caption{T-SNE~\cite{maaten2008visualizing} visualization of feature embeddings learned from various KD methods over randomly sampled 10 classes from CIFAR-100. Moreover, we use Recall@1 to quantify intra-class variations of embedding space. Recall@$k$ denotes the percentage of test images that have at least one image from the same class in $k$ nearest neighbours in the embedding space.} 
	\label{visual} 
\end{figure*} 
\subsection{Effect of four different knowledge forms for offline HSSAKD} In Section ~\ref{training_single}, we have examined the effectiveness of four different auxiliary tasks for training a single network. To further verify the efficacy, we employ four different distributions referred in Section~\ref{C} derived from these auxiliary tasks as knowledge to conduct offline KD. As shown Table~\ref{offline}, distilling our proposed self-supervision augmented distribution as auxiliary knowledge can significantly enhance performance and outperform other alternative distributions by large margins. Compared with others, it is demonstrated as richer knowledge that encodes more meaningful information derived from the joint task of supervised and self-supervised tasks.  

\begin{table}
	\caption{Top-1 accuracy (\%) comparison towards various self-supervised pretext tasks for \emph{offline HSSAKD} on the student networks of WRN-16-2 and ShuffleNetV1 supervised through the pre-trained teacher network WRN-40-2 on CIFAR-100.}
	
	\centering
		\begin{tabular}{lcccccccccccc}
			\toprule
			Self-supervised pretext task & WRN-16-2 &ShuffleNetV1  \\
			\midrule
			6-way color  channal permutation & 75.28$_{(\pm 0.16)}$  &77.05$_{(\pm 0.26)}$ \\
			4-way jigsaw puzzles & 77.15$_{(\pm 0.24)}$  &79.23$_{(\pm 0.35)}$ \\
			4-way random rotations & \textbf{78.67}$_{(\pm 0.20)}$ & \textbf{80.11}$_{(\pm 0.32)}$ \\
			\bottomrule
	\end{tabular}
\label{pretext}
\end{table}
\subsection{Effect of various self-supervised auxiliary tasks} In this paper, we investigate three self-supervised classification tasks over the input image: 
\begin{itemize}
	\item \emph{6-way color  channal permutation}: \{RGB, RBG, GRB, GBR, BRG, BGR\}
	\item  \emph{4-way jigsaw puzzles}: Jigsaw splits the input image into $2\times 2=4$ non-overlapping patches. We can rearrange these 4 input patches and lead to $4!=24$ possible permutations. We re-organize these patches into the original image format and forces the network to classify the correct permutation. To shrink the class space, we select 4 permutations with maximum Hamming distance following~\cite{noroozi2016unsupervised}.
	\item \emph{4-way random rotations}: $\{0^{\circ},90^{\circ},180^{\circ},270^{\circ}\}$
\end{itemize}
 In theory, different self-supervised tasks would lead to different qualities of representation learning. We examine these three self-supervised tasks to construct self-supervision augmented distributions for offline HSSAKD, respectively. As shown in Table~\ref{pretext}, we find that 4-way random rotation is a more favourable self-supervised pretext task. The observation is consistent with the previous SSKD~\cite{DBLP:conf/eccv/XuLLL20}.
\subsection{Effect of the temperature $\tau$} The temperature $\tau$ is a common hyper-parameter in KD to smooth the produced probability distributions. In theory, a larger $\tau$ would result in a smoother distribution. To search the best $\tau$, we vary it from 1 to 5, which is a reasonable range verified by previous KD works~\cite{hinton2015distilling,chen2021cross}. As shown in Table~\ref{temperature}, we find that HSSAKD is robust to the temperature $\tau$ and $\tau=3$ may be a more suitable choice.
\begin{table}
	\caption{Top-1 accuracy (\%) comparison towards ablation study of temperature $T$ on the student networks of WRN-16-2 and ShuffleNetV1 supervised through the pre-trained teacher network WRN-40-2 for \emph{offline HSSAKD} on CIFAR-100.}
	\centering
	\begin{tabular}{cccccc}
		\toprule
		Temperature $T$& $T=1$ &$T=2$ &$T=3$ &$T=4$ &$T=5$  \\
		\midrule
		WRN-16-2 &78.20 & 78.35&\textbf{78.67} &78.52 & 78.54\\
		ShuffleNetV1 &78.20 & 79.65 &\textbf{80.11} &80.03 &80.01 \\
		\bottomrule
	\end{tabular}
	\label{temperature}
\end{table}

 \begin{figure}[tbp]  
	\centering  
	\includegraphics[width=1\linewidth]{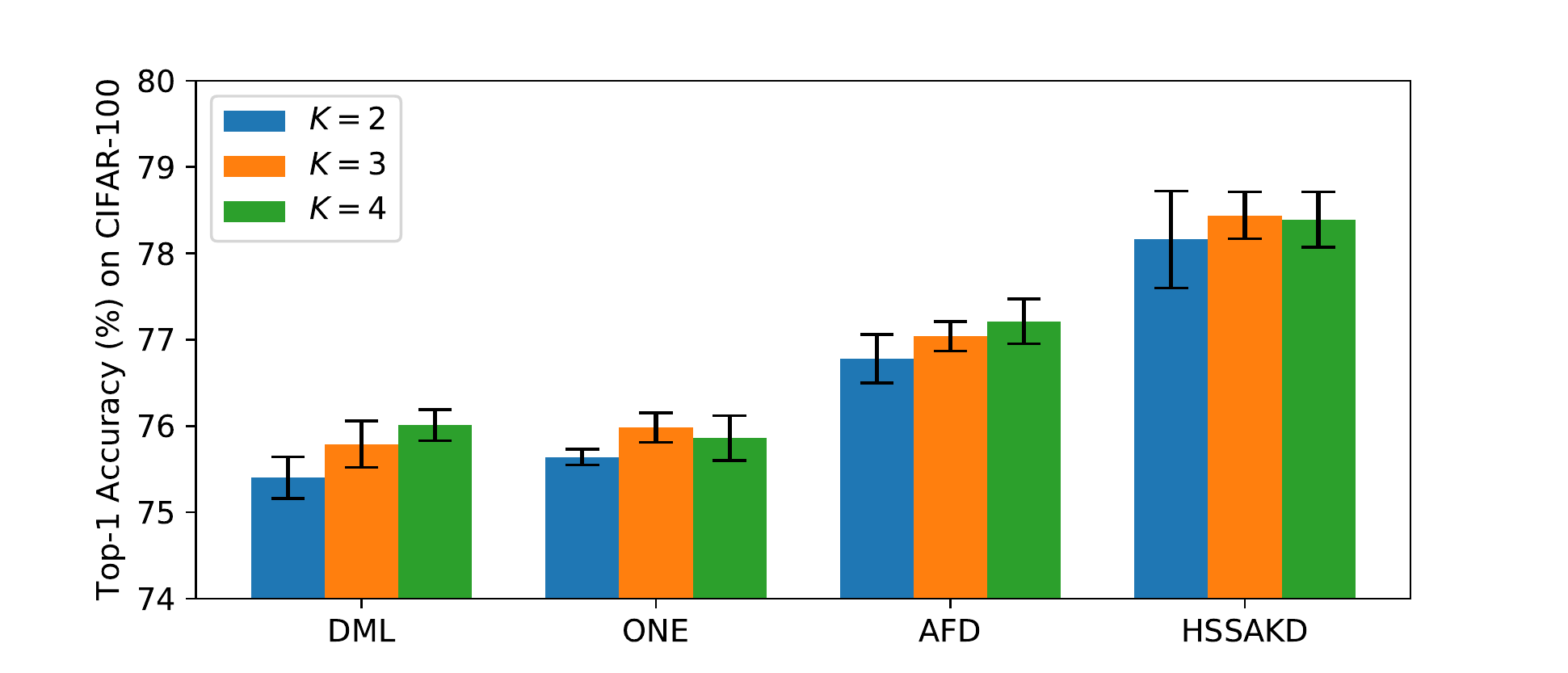}
	\caption{Impact of the number of networks $K$ for online KD.}  
	\label{K}
\end{figure}

\subsection{Impact of the number of networks $K$ for online KD} We investigate the performance of multi-student based online KD as the number of networks $K$ increases. Fig.~\ref{K} plots the accuracy comparison among various online KD methods with the number of networks $K$ changing from 2 to 4.  We find that accuracy gains generally increase from $K=2$ to $K=3$ but saturate at $K=4$. Moreover, even in the case of the minimum training capacity of $K=2$, our HSSAKD can still substantially beat other compared online KD methods with $K=4$ networks. The results demonstrate the superiority of our self-supervision augmented distribution as an effective knowledge form.

\subsection{Visualization of learned feature embeddings.}
As shown in Fig.~\ref{visual}, we use t-SNE~\cite{maaten2008visualizing} to visualize pooled feature embeddings after the penultimate layer and report quantitative results on Recall@1.  We can observe that our HSSAKD achieves higher visual separability and Recall@1 than other KD methods. The results demonstrate the efficacy of HSSAKD for generating a more discriminative feature space via distilling hierarchical self-supervised augmented knowledge. The better feature space further leads to a significant improvement of recognition accuracy.

\subsection{Analysis of Training Overhead.}
The offline KD conducts a two-stage pipeline that trains a teacher network and a student network sequentially. For training overhead, we compare our HSSAKD with the vanilla KD~\cite{hinton2015distilling}. Compared with KD, extra training costs of our HSSAKD lies in attached auxiliary branches for learning and distillation. We take the widely used WRN-40-2-WRN-40-1 as a teacher-student pair for illustration. For the teacher, computation of the original WRN-40-2 is 330MFLOPs. After appending auxiliary branches, the computation increases to 770MFLOPs. For the student, computations of the original WRN-40-1 and its auxiliary branches augmented counterpart are 80MFLOPs and 190MFLOPs, respectively. In practice, the overall pipeline training time of our HSSAKD is about $2\times$ than the vanilla KD. However, HSSAKD outperforms the vanilla KD with a significant 3.1\% accuracy gain, demonstrating the rationality of training costs.

 \section{Conclusion}
We propose a self-supervision augmented task for KD and further transfer such rich knowledge derived from hierarchical feature maps leveraging well-designed auxiliary branches. Our method achieves SOTA performance on the standard image classification benchmarks than other offline and online KD methods. It is also demonstrated that HSSAKD can guide the network to learn well-general feature representations for semantic recognition tasks due to the well-combined self-supervised auxiliary task. Due to the excellent performance, we believe that our HSSAKD may attract wide attention in the community of KD. In the future, we will apply our method to compress some networks for other visual recognition tasks, for example, object detection~\cite{chen2017learning} and semantic segmentation~\cite{yang2022cross}.


%
%
%
%
%
%

\ifCLASSOPTIONcaptionsoff
  \newpage
\fi



%
\bibliographystyle{IEEEtran}
\bibliography{IEEEbib}

\begin{IEEEbiography}[{\includegraphics[width=1in,height=1.25in,clip,keepaspectratio]{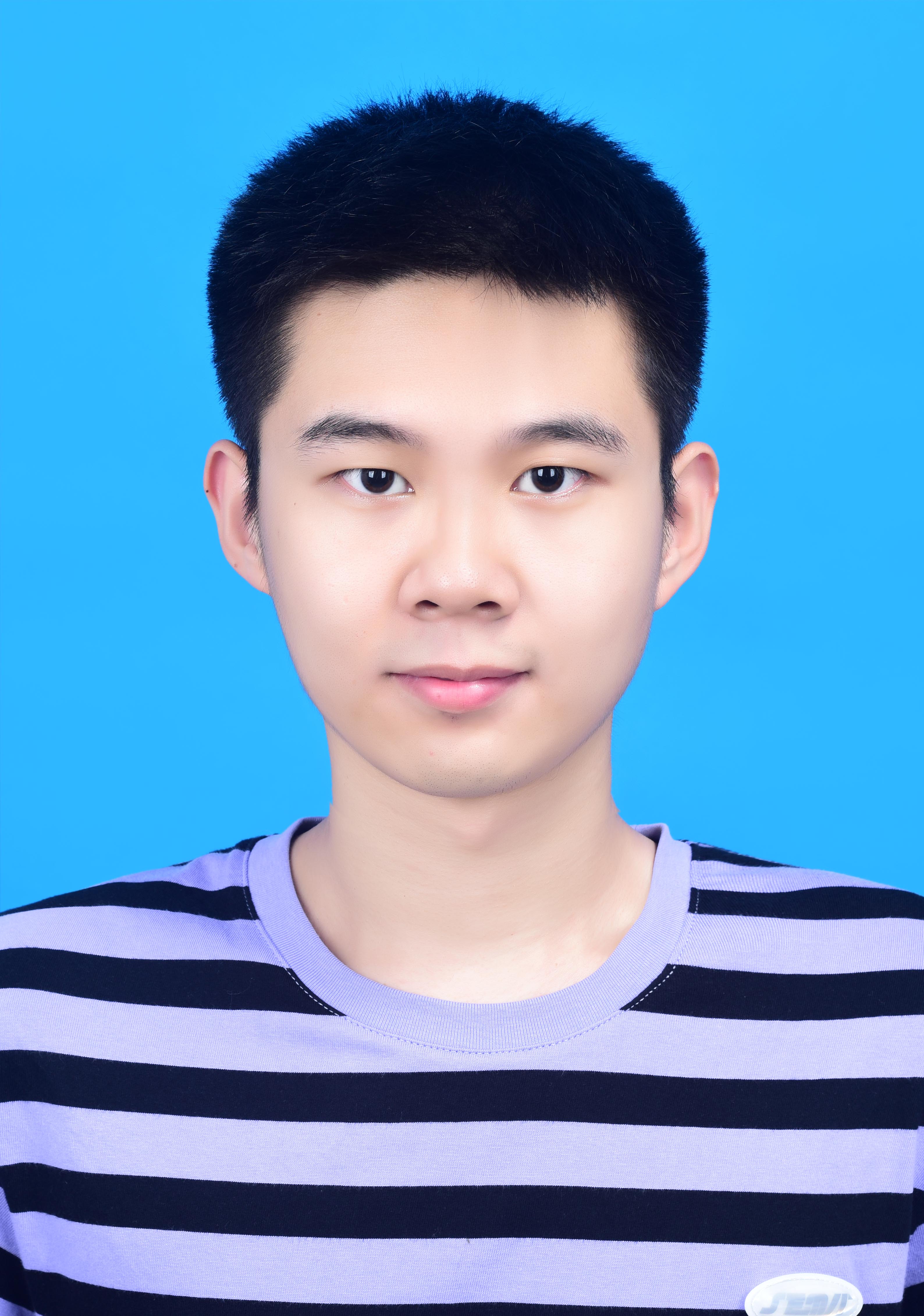}}]{Chuanguang Yang}
	received the B.Eng. degree from Shandong Normal University, Jinan, China, in 2018. He is currently pursuing a Ph.D. degree with the Institute of Computing Technology, Chinese Academy of
	Sciences, China. His research interests include knowledge distillation, visual representation learning and image classification.
\end{IEEEbiography}

\begin{IEEEbiography}[{\includegraphics[width=1in,height=1.25in,clip,keepaspectratio]{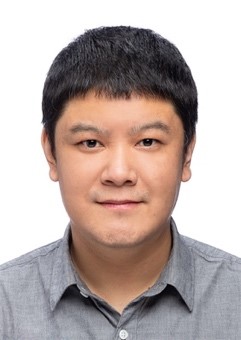}}]{Zhulin An}
	received the B.Eng. and M.Eng. degrees in computer science from Hefei University of Technology, Hefei, China, in 2003 and 2006, respectively and the Ph.D. degree from the Chinese Academy of Sciences, Beijing, China, in 2010.
	He is currently with the Institute of Computing Technology, Chinese Academy of Sciences, where he became a Senior Engineer in 2014. His current research interests include optimization of deep neural network and lifelong learning.
\end{IEEEbiography}

\begin{IEEEbiography}[{\includegraphics[width=1in,height=1.25in,clip,keepaspectratio]{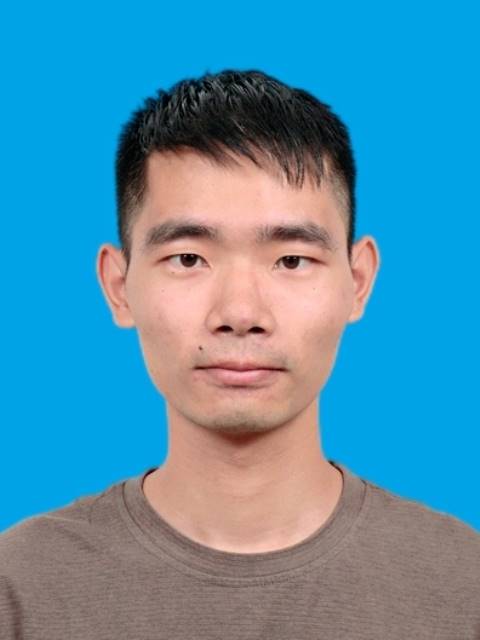}}]{Linhang Cai}
	received the B.Eng. degree from Sun Yat-sen University, Guangzhou, China, in 2019 and the M.Eng. degree with the Institute of Computing Technology, Chinese Academy of
	Sciences, Beijing, China, in 2022. His research interests include knowledge distillation, network pruning and image classification.
\end{IEEEbiography}

\begin{IEEEbiography}[{\includegraphics[width=1in,height=1.25in,clip,keepaspectratio]{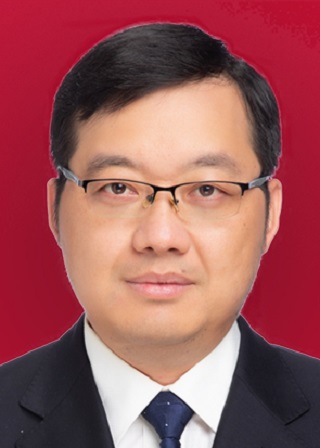}}]{Yongjun Xu}
	is a professor at Institute of Computing Technology, Chinese Academy of Sciences (ICT-CAS) in Beijing, China. He received his B.Eng. and Ph.D. degree in computer communication from Xi'an Institute of Posts \& Telecoms (China) in 2001 and Institute of Computing Technology, Chinese Academy of Sciences,
	Beijing, China in 2006, respectively. His current
	research interests include artificial intelligence
	systems, and big data processing.
	
\end{IEEEbiography}

%

%
%
%




\end{document}